\documentclass[10pt,journal]{IEEEtran}
\IEEEoverridecommandlockouts         
\PassOptionsToPackage{table,dvipsnames,svgnames}{xcolor}
\usepackage{xcolor}                  

\usepackage[colorlinks=true,
            linkcolor=blue,
            citecolor=blue,
            urlcolor =blue]{hyperref}

\usepackage{orcidlink}            
\usepackage{cite}
\usepackage{amsmath,amssymb,amsfonts,amsthm}
\usepackage{upgreek}
\usepackage{algorithmic}
\usepackage{graphicx}
\usepackage{textcomp}
\usepackage{subcaption}
\usepackage{algorithm}
\usepackage{booktabs}
\usepackage{placeins}
\usepackage{fancyhdr}
\usepackage{tikz}
\usepackage{tikz-3dplot}
\usepackage{pgfplots}
\pgfplotsset{compat=1.18}
\usetikzlibrary{arrows.meta,positioning,calc,patterns}

\colorlet{body}{blue!45}
\colorlet{edge}{blue!70}

\tikzset{
    >=Stealth,  font=\small,
    module/.style = {draw=gray, rounded corners=2pt, fill=gray!15, minimum width=28mm, minimum height=6mm},
    light/.style   = {->, line width=1.2pt, draw=LimeGreen!100, dashed},
    flow/.style   = {->, line width=1.2pt, draw=blue!50},
    back/.style   = {->, line width=1.2pt, draw=orange!80},
    tire/.style   = {draw, rounded corners=0.3pt, fill=darkgray!80, width=5mm, height=10mm},
       robot/.pic = {
      \coordinate (-center) at (0,0);          
      \draw[fill=blue!20] (-0.5,-0.2) rectangle ++(1,0.4);   
      \draw[fill=black] (-0.35,-0.35) circle (0.12);         
      \draw[fill=black] ( 0.35,-0.35) circle (0.12);
      \draw[->, very thick, red] (0,0.25)--++(0,0.5);        
  }, 
    tree/.pic = {
    \begin{scope}[rotate=-90] 
      \fill[RawSienna!20] (0,0) -- (30:.3) arc (30:-30:.3) -- cycle;
      \draw[RawSienna!50] (0,0)--(30:.3) (0,0)--(-30:.3);
    \end{scope}
    \begin{scope}[yshift=0.9cm, rotate=-90]
      \fill[ForestGreen!20] (0,0) -- (20:1) arc (20:-20:1) -- cycle;
      \draw[YellowGreen!50] (0,0)--(20:1) (0,0)--(-20:1);
    \end{scope}
    \coordinate (-center) at (0,0);
  }, 
    rzr/.pic = {
    \node[inner sep=0pt, anchor=center] (-img)
          {\includegraphics[width=2.5cm]{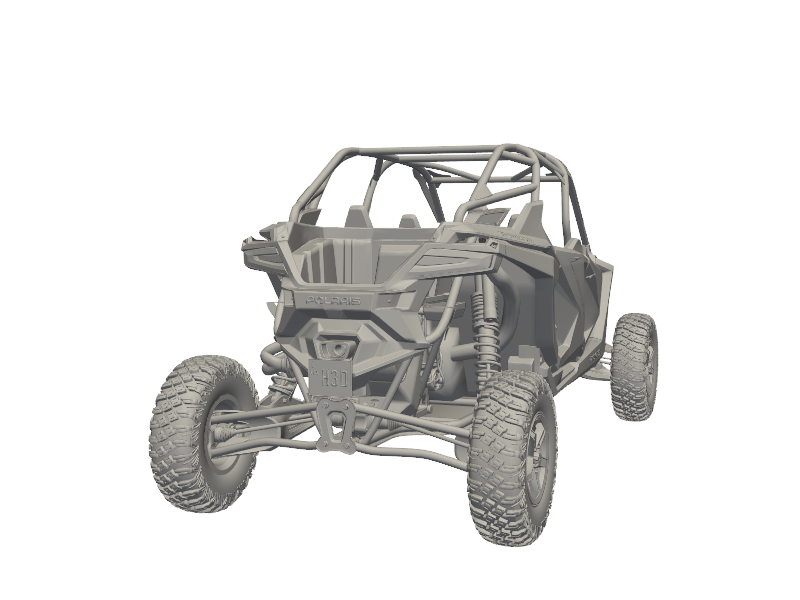}};
    \coordinate (-center) at (-img.center);
    \coordinate (-south) at (-img.south);
  },
    partial ellipse/.style args={#1:#2:#3}{
        insert path={+ (#1:#3) arc (#1:#2:#3)}
    },
    perception/.pic = {
    \begin{scope}         
      \path[fill=blue!10, opacity=.5, draw=blue!20]
            (0,0)                             
            -- +(30:3cm and 0.65cm)          
            arc  (30:100:3cm and 0.65cm)     
            -- cycle;                          
    \end{scope}
    \coordinate (-north) at (0.0,0.5);
    },
      traj/.pic = {%
    \pgfkeys{/traj/.cd,
      length/.initial = 3.0,     
      height/.initial = 0.9,    
      bend/.initial   = 0.6,    
      color/.initial  = blue!60, 
      style/.initial  = solid,   
      opacity/.initial= 0.9,     
    }
    \pgfkeysgetvalue{/traj/length}{\L}
    \pgfkeysgetvalue{/traj/height}{\H}
    \pgfkeysgetvalue{/traj/bend}{\B}
    \pgfkeysgetvalue{/traj/color}{\C}
    \pgfkeysgetvalue{/traj/style}{\S}
    \pgfkeysgetvalue{/traj/opacity}{\O}
   
    \pgfmathsetmacro{\xC}{\L*\B}       
    \pgfmathsetmacro{\yC}{\H*0.6}       
    
    \draw[->, line width=1pt, draw=\C, \S, opacity=\O]
          (0,0) .. controls (\xC,\yC) and (\L-\xC,\H-\yC) .. (\L,\H);
  }
}

\def\BibTeX{{\rm B\kern-.05em{\sc i\kern-.025em b}\kern-.08em
    T\kern-.1667em\lower.7ex\hbox{E}\kern-.125emX}}

\DeclareMathOperator*{\argmin}{arg\,min}

\newtheorem*{remark}{Remark} 
\newcommand{\orcidSup}[1]{\raisebox{.25em}{\orcidlink{#1}}}

\begin{document}

\title{Implicit Dual-Control for Visibility-Aware Navigation Unstructured Environments}

\author{Benjamin Johnson$^{1,}$\orcidSup{0009-0001-5388-0545}, Qilun Zhu$^{1,}$\orcidSup{0000-0003-2995-2470} , Robert Prucka$^{1,}$\orcidSup{0000-0001-5958-1565}, \\
        Miriam A. Figueroa-Santos$^{2,}$\orcidSup{0000-0001-5553-6652}, Morgan J. Barron$^{2,}$\orcidSup{0009-0001-0432-3611}, and Matthew P. Castanier$^{2,}$\orcidSup{0000-0002-3646-382X}%
\thanks{Acknowledgment: This work was supported by Clemson University’s Virtual Prototyping of Autonomy Enabled Ground Systems (VIPR-GS), a US Army Center of Excellence for modeling and simulation of ground vehicles, under Cooperative Agreement W56HZV-21-2-0001 with the US Army DEVCOM Ground Vehicle Systems Center (GVSC). DISTRIBUTION STATEMENT A. Approved for public release; distribution is unlimited. OPSEC9799}%
\thanks{$^1$B. Johnson, Q. Zhu, and R. Prucka are with the Department of Automotive Engineering, Clemson University, Greenville, SC, USA (e-mail:\tt\footnotesize{ \{bij, qilun, rprucka\}@clemson.edu)}.}%
\thanks{$^2$M. A. Figueroa-Santos, M. J. Barron, and M. P. Castanier are with the Vehicle Performance and Analytic Development M\&S Team, Ground Vehicle Systems Center (GVSC), Warren, MI, USA (e-mail:\tt\footnotesize{ \{miriam.a.figueroa-santos, morgan.j.barron4, matthew.p.castanier\}.civ@army.mil)}.}
\thanks{Corresponding Author: B. Johnson \tt\footnotesize{bij@clemson.edu}}
}

\maketitle
\thispagestyle{fancy}
\pagestyle{fancy}

\begin{abstract}
Navigating complex, cluttered, and unstructured environments that are a priori unknown presents significant challenges for autonomous ground vehicles, particularly when operating with a limited field of view(FOV) resulting in frequent occlusion and unobserved space. This paper introduces a novel visibility-aware model predictive path integral framework(VA-MPPI). Formulated as a dual control problem where perceptual uncertainties and control decisions are intertwined, it reasons over perception uncertainty evolution within a unified planning and control pipeline. Unlike traditional methods that rely on explicit uncertainty objectives, the VA-MPPI controller implicitly balances exploration and exploitation, reducing uncertainty only when system performance would be increased. The VA-MPPI framework is evaluated in simulation against deterministic and prescient controllers across multiple scenarios, including a cluttered urban alleyway and an occluded off-road environment. The results demonstrate that VA-MPPI significantly improves safety by reducing collision with unseen obstacles while maintaining competitive performance. For example, in the off-road scenario with 400 control samples, the VA-MPPI controller achieved a success rate of 84\%, compared to only 8\% for the deterministic controller, with all VA-MPPI failures arising from unmet stopping criteria rather than collisions. Furthermore, the controller implicitly avoids unobserved space, improving safety without explicit directives. The proposed framework highlights the potential for robust, visibility-aware navigation in unstructured and occluded environments, paving the way for future advancements in autonomous ground vehicle systems.\end{abstract}

\section{Introduction}
\begin{figure}[ht]
    \centering
    \hspace*{0cm}
    \includegraphics[width=1.0\linewidth]{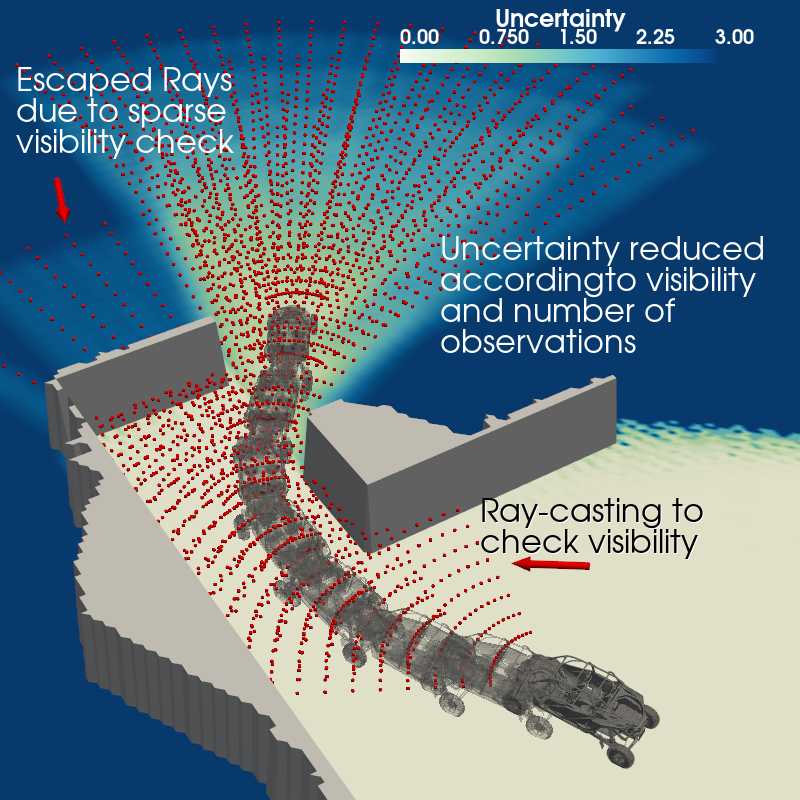}
    \caption{Visibility-Aware MPPI allows planning into uncertain areas by reducing uncertainty in the prediction horizon through a ray-casting model. Due to the sparse visibility check along each ray, we occasionally get an "escaped" ray. A denser ray would rectify the issue, but we did not find it to be worth the extra computational load.}
    \label{va_mppi_uncertainty_update}
\end{figure}

This paper investigates the challenge of navigating complex, cluttered, and unstructured environments (e.g., off-road or off-trail) which are a priori unknown. A visibility-aware implicit dual control model predictive path integral(VA-MPPI) framework (see Fig.~\ref{va_mppi_uncertainty_update}) is presented to quantify and reason about the evolution of perception uncertainties in a unified planning and control pipeline. Particular focus is placed on addressing uncertainties in unobserved spaces caused by occlusions or field-of-view(FOV) limitations. 
In the dual control paradigm, a controller works to balance exploration(probing the environment to reduce uncertainty) and exploitation(achieving the control objective)~\cite{FELDBAUM1963541}. A system is said to exhibit the dual effect when control inputs impact both the system states and the uncertainty of the states~\cite{bar_shalom_1100635}.  Robot vision is particularly well-suited to dual control due to its parallels with human perception which is inherently active and probing. We do not just passively observe but actively look around, particularly in new environments~\cite{active_perception}.

Conceptually, dual control is very intuitive. Humans frequently navigate new and unknown environments by continuously quantifying and weighting uncertainties in a calculated decision making process that is both adaptive and instinctive. As such, aggressive behavior tends to emerge under conditions of low uncertainty, while high uncertainty prompts more conservative actions. In addition, because decisions influence the evolution of these uncertainties, a natural balance between uncertainty reduction and task progress emerges. For a robot operating in an unknown environment, similar behavior is desirable. However, instilling this capability implicitly remains a challenging and open research question. 

The difficulty intensifies in environments lacking preexisting maps, discrete lanes, or markings that guide decisions. In addition, off-road and off-trail environments are often more rugged or cluttered, with occlusions caused by large obstacles or undulating terrain. Coupled with limited field of view(FOV), these factors pose significant challenges to navigation. In the absence of a priori knowledge, uncertainty in the environment is unavoidable. Successful navigation in such conditions necessitates incorporating this uncertainty into the planning stage.
\subsection{Map Representation}
In the traditional motion planning paradigm, a binary approach is taken marking space as free(safe) or occupied(unsafe) \cite{tordesillas2021fasterfastsafetrajectory, ramp}. This results in unknown space falling into one of these two categories, each with a drastically different effect on the planner. Treating the unknown space as occupied yields safe, but often overly conservative trajectories. While treating it as free leads to more aggressive trajectories, but risks collision with unseen obstacles.  There have been a number of attempts at bridging this dichotomy and improve the handling of unknown or uncertain space. 

A number of works directly handle the uncertainty in the planning process by either considering known and unknown separately~\cite{tordesillas2021fasterfastsafetrajectory, ramp}, or taking a probabilistic approach~\cite{isitworthreasoning, wang2023multiplehypothesispathplanninguncertain, probabilistic_traversability, risk_aware_off_road_navigation, fan2021stepstochastictraversabilityevaluation, Fan_2022}. Sharma et. al.~\cite{ramp} plan paths through known and unknown space by limiting the vehicle to different top speeds in the two regions, this forces the vehicle to slow down in regions of high uncertainty, but allows the vehicle to move faster in regions that are known. Tordesillas et. al.~\cite{tordesillas2021fasterfastsafetrajectory} plan two different trajectories, one which prioritizes speed even in unknown space, and another that always has a safe stopping distance within the known region. The first segment is the same for both trajectories so should a feasible solution not be available in the next time step the vehicle can safely come to a stop. 

Still others take a probabilistic approach to handling the uncertainty. Banfi et. al.~\cite{isitworthreasoning} plan an initial path hypothesis to goal, should the path remain in known space they proceed to execute. If the first hypothesis enters unknown space, a second path is planned where the first segment of each hypothesis is the same and ends with a next best view pose. Wang et. al.~\cite{wang2023multiplehypothesispathplanninguncertain} consider the uncertainty of unobserved or poorly observed obstacles while using a graph based multi-hypothesis planner. As obstacle detections become more certain, unsafe hypothesis are trimmed away. 

A significant number of works have moved away from simply modeling occupancy and began using probabilistic cost maps which consider not just the risk of collision, but also dynamic and efficiency constraints on traversability~\cite{probabilistic_traversability, risk_aware_off_road_navigation, fan2021stepstochastictraversabilityevaluation, Fan_2022, castro2023doesfeelselfsupervisedcostmap,triest2023learningriskawarecostmapsinverse} 
Fan et. al.~\cite{fan2021stepstochastictraversabilityevaluation} use a 2.5D elevation map to generate an ensemble of risk maps based on various risky behaviors(roll over, collision, slippage, etc..), the work is extended to a completely learned risk map in~\cite{Fan_2022}. These maps are then used in a downstream planner showing improved saftey.

A common approach is to learn traversability by combining geometric and propriotceptive information to determine factors like achievable speed~\cite{risk_aware_off_road_navigation}, traction distributions~\cite{probabilistic_traversability}, and ride comfort~\cite{castro2023doesfeelselfsupervisedcostmap}. Triest et. al.\cite{triest2023learningriskawarecostmapsinverse} employs inverse reinforcement learning to learn risk-aware cost maps from expert demonstration.  However, all of these works explicitly avoid or assign high cost to unknown or unmapped regions when planning.  On the other-hand, VA-MPPI does not explicitly avoid unobserved space, rather we recognize that entering such space will lead to an observation and thus implicitly reduce uncertainty. 

In an effort to avoid having unknown space in the map many works focus on data driven inpainting to build more complete and usable maps~\cite{planning_paths_by_imagining, han2022planningpathsocclusionsurban, learning_3d_occupancy, terrain_net, reconstructing_occluded, triest2024unrealnetlearninguncertaintyawarenavigation} or active exploration\cite{georgakis2022uncertaintydrivenplannerexplorationnavigation, baek2025pipeplannerpathwiseinformation, uncertainty_aware, coslam, niceslam, naruto}.

Inpainting has a rich history in computer vision, but modern data driven techniques have shown great promise in improving robotic navigation.  By removing the unknown space in the map,  planners can plan into areas that would traditionally be avoided.  This approach has proven to improve planning by enabling longer and smoother paths closer to ground truth data~\cite{planning_paths_by_imagining, han2022planningpathsocclusionsurban, learning_3d_occupancy, terrain_net, reconstructing_occluded, triest2024unrealnetlearninguncertaintyawarenavigation}. Han et. al.~\cite{planning_paths_by_imagining} train a neural network to inpaint occluded birds-eye view semantic lidar data showing an improvement in path planning. Their work was extended in~\cite{han2022planningpathsocclusionsurban} to include self-supervised labeling and vehicle dynamics during planning. St{\"o}ze et. al. generate pseudo occlusions from known regions of the map to true occlusions in contextually consistent approach. Wang et. al.~\cite{learning_3d_occupancy} extend the inpainting idea to handle full 3D environment, but is limited to small indoor environments. For large scale off-road environments,~\cite{terrain_net} models terrain using a multi-layer 2.5D representation, inpainting each layer before fusing them into a dense occlusion free cost map for MPPI planning. Typically these methods do not account for the uncertainty of the inpainted map, which could lead to over confidence in some scenarios. Triest et al. \cite{triest2024unrealnetlearninguncertaintyawarenavigation} do account for the uncertainty by learning an uncertainty aware traversability map which implicitly learns to inpaint missing regions, but they require dense a priori data of their training environments. While inpainting demonstrates improvements over baseline maps, it is data-intensive and struggles to generalize to novel, unseen environments, making it prone to errors when navigating outside its training set. VA-MPPI however, employs inpainting through a Telea inpainting scheme which does not require learned priors. Uncertainty is handled through a separate uncertainty layer which is only reduced through observation. 

Active mapping~\cite{georgakis2022uncertaintydrivenplannerexplorationnavigation, baek2025pipeplannerpathwiseinformation, uncertainty_aware, coslam, niceslam, naruto}focuses on generating navigable representations, such as occupancy grids or topological maps, with an emphasis on exploration to maximize coverage and reduce map uncertainty. This often involves selecting the next-best view for gathering perception data. Works like~\cite{uncertainty_aware, georgakis2022uncertaintydrivenplannerexplorationnavigation, baek2025pipeplannerpathwiseinformation} combine ensemble inpainting to quantify map uncertainty, targeting high-uncertainty regions for further exploration. Others leverage neural implicit or hybrid representations to generate detailed 3D maps~\cite{coslam, niceslam, naruto}. These works use ray-marching to evaluate uncertainty in the map and plan next a next best view. This served as an inspiration to  VA-MPPI which uses ray-casting to evaluate visibility and thus reduce uncertainty. While these methods enhance mapping and reconstruction capabilities, they predominantly address dense mapping tasks which attempts to reduce all uncertainty. VA-MPPI on the other hand reduces uncertainty only in regions of the map necessary for fast point to point navigation.
\subsection{Perception-Aware Planning}
Rather than directly incorporating the uncertainty into the planner, many works attempt to mitigate the uncertainty entirely through explicit perception objectives. Perception-aware and visibility-aware planning has been extensively studied, particularly for UAVs. While often not explicitly formulated as such, these approaches are inherently working to solve the dual problem as actions are taken to achieve some vision objective which in turn reduces uncertainty. The general approach involves formulating an explicit perception objective in the cost function to minimize uncertainty or maximize information gain. In contrast, the proposed framework does not rely on an explicit uncertainty or vision objective. 

A vast body of work focuses on visual state estimation~\cite{pa_receding_horizon, perceptionawarepathplanning, online_optimal_perception_aware, chen2024apaceagileperceptionawaretrajectory, pa_traj_generation_aggressive_quadrotor, fisherinformationfield, probabilisticuncertaintyquantificationprediction} during navigation. These approaches prioritize trajectories which keep specific landmarks or rich visual information in the FOV to minimize localization and pose uncertainties. Zhang et. al.~\cite{pa_receding_horizon} employ monocular odometry for mapping and state estimation while optimizing perceptual quality and collision avoidance. Salaris et. al. ~\cite{online_optimal_perception_aware} take minimizing the smallest eigenvalue of the constructability Grammian along a B-spline trajectory as a measure for information gain. Costante et. al.~\cite{perceptionawarepathplanning} incorporate dense photometric information, optimizing directly over pixel intensities instead of feature-based matching. Chen et. al.~\cite{chen2024apaceagileperceptionawaretrajectory} introduce a differentiable visibility model which enhances the number of matched features between consecutive frames while minimizing their parallax angle. Murali et. al.~\cite{pa_traj_generation_aggressive_quadrotor} focuses on generating aggressive trajectories while simultaneously considering co-visibility of features. Chen et. al.~\cite{probabilisticuncertaintyquantificationprediction} develop an uncertainty prediction neural network model for visual localization , which uses the number of keypoint matches as a metric for uncertainty quantification. 

Beyond localization, several works optimize visibility of a(potentially occluded or non stationary) point of interest (POI), balancing this with speed, smoothness, feasibility, and safety~\cite{Raptor, svpto, vision_based_aggressive_tracking, Panther, pampc, visibility_aware_tracking, perception_limited_mpc}. Cascaded approaches such as~\cite{Raptor, svpto, Panther} start with a topological planner and refine trajectories. Zhou et. al.~\cite{Raptor} use risk-aware replanning to ensure safe distances and visibility of hazardous regions, followed by yaw optimization for information gain. Similarly,~\cite{svpto, Panther} jointly optimize speed, safety, and visibility of a target or dynamic obstacles. Penin et. al.~\cite{vision_based_aggressive_tracking} employ a multi-objective optimization to initialize trajectories, followed by fast SQP-based re-planning for dynamic scenarios.

Joint optimization methods address the potential sub-optimality of separate objectives~\cite{pampc, visibility_aware_tracking, perception_limited_mpc, cross_entropy, pixelmpc}. Falanga et. al.~\cite{pampc} maintain a POI in the FOV while considering under actuation and actuator limits. Wang et. al.~\cite{visibility_aware_tracking} optimize trajectories based on observation distance, angle, and occlusion while tracking a quadrotor. Lu et. al. ~\cite{perception_limited_mpc} utilize sampling-based MPC to jointly optimize motion and perception constraints, enabling safe navigation within visible regions while maintaining localization of a POI. Learning-based methods such as~\cite{cross_entropy, pixelmpc} integrate perception models directly into planning. Masnavi et. al.~\cite{cross_entropy} learn occlusion models from raw LiDAR data, while Lee et. al.~\cite{pixelmpc} use deep optical flow networks to predict pixel dynamics, demonstrated in drone racing scenarios.

Most of these systems focus on UAVs and often exploit their differential flatness and ability to independently optimize for yaw and trajectory directions. Ground vehicles, constrained to operate on surfaces and lacking independent yaw control, present unique challenges for perception-aware planning. Limited work has addressed this for ground vehicles~\cite{probabalistic_visibility_aware, negotiating_visibility}. Gao et. al.~\cite{probabalistic_visibility_aware} model target visibility probabilistically, but assume a priori map knowledge and operate at low speeds ($<$1 m/s). Higgins and Bezzo~\cite{negotiating_visibility} develop a geometry based vision objective assuming an a priori known map and optimal guiding trajectories, to force better visibility around occluded corners.

Visibility-aware MPPI builds on these contributions by addressing real-time navigation challenges in cluttered, occluded environments. Unlike the vast body of perception-aware literature however, it implicitly handles perception uncertainties without relying on explicit vision objectives, enabling robust and dynamic navigation.
\subsection{Dual Control}
Dual control has been explored in both explicit and implicit contexts~\cite{mesbah_dual_survey, mesbah_pa, active_learning_dual_control, safe_motion_planning_optimality_benchmarks, GHEZZI2023100841_steel}. Explicit dual control methods integrate active probing into the control framework, often through heuristic-based measures or direct incorporation of uncertainty-reduction objectives. These methods include persistent excitation techniques and multi-objective formulations, such as augmenting standard MPC cost functions with terms that reward parameter uncertainty reduction. Implicit dual control, on the other hand, approximates the stochastic dynamic programming problem to indirectly account for the dual-control effect~\cite{mesbah_dual_survey}. 

Recent works have extended dual-control principles into uncertainty-aware navigation. Janson et al.~\cite{safe_motion_planning_optimality_benchmarks} demonstrates how evaluating trajectories against Inevitable Collision States (ICS) implicitly balances exploration and exploitation. By dynamically adapting trajectories to reveal unobserved areas while ensuring safety, this work aligns closely with dual-control principles, even though it does not explicitly frame itself in this context.  Additionally, they do not assume limited vision(range, fov, etc) and  the planner is very slow (0.5-1s). Bonzanini et at.~\cite{mesbah_pa} propose a perception-aware MPC which is formulated as a dual control problem. They utilize the uncertainty of environment estimate, which is a direct result of the perception system, to adjust the tightening of chance constraints. However, it is formulated as an explicit approach, which always seeks to drive uncertainty down even if it does not affect the performance. Additionally their use of a gradient based solver can impose performance constraints due to its sequenctial nature and the non-convexity of trajectory planning.  Our VA-MPPI on the other-hand is formulated implicitly thus only we only reduce uncertainty when it affects system performance. MPPI is also a gradient free solver and highly parallelizable. Knaup et al.~\cite{active_learning_dual_control} also employs MPPI in an implicit dual-control framework from which we take inspiration, however it is formulated for interaction-aware navigation for learning other drivers behavior.
\subsection{Contributions}
Based on the preceding literature review, this paper addresses a critical gap in research by proposing a novel approach to visibility-aware planning and control for off-road, non-holonomic vehicles. Specifically, our contributions are as follows:
\begin{enumerate}
    \item Implicit Dual-Control Framework: We present the first implementation of visibility-aware planning and control as an implicit dual-control problem. Instead of explicitly minimizing uncertainty, our method incorporates it naturally into the decision-making process, reducing uncertainty only when it directly impacts system performance.
    \item Adaptation to full size Off-Road Vehicles: We extend visibility-aware planning to ground vehicles with limited FOV, constrained to operate in off-road environments. This marks a significant departure from prior works focused predominantly on aerial vehicles and slow skid steer robots.
    \item Demonstration in Unknown Environments: Through simulation, we demonstrate the feasibility, effectiveness, and real-time ability of our approach for navigating heavily occluded and previously unseen environments. The proposed controller shows promise for real-world applications where robust, real-time, visibility-aware navigation is critical.
\end{enumerate}
\section{Visibility-Aware Implicit-Dual Control Problem Formulation}
\begin{figure}
    \centering
        \begin{tikzpicture}
        \node[draw=black!40, dashed, rounded corners=2pt, fill=YellowGreen!15, opacity=0.2,
              minimum width=45mm, minimum height=45mm,
              anchor=north west] (world) at (-5mm,0) {};
        \node[font=\small] at (world.north) [yshift=-8pt] {Unknown Terrain};      
        \node[draw=SpringGreen!80, rounded corners=2pt, fill=SpringGreen!15, minimum width = 35mm, minimum height=8mm, right=5mm of world.north east, anchor=north west]
              (mapping) {Perception / Mapping};
        \node[draw=Purple!80, rounded corners=2pt, fill=Purple!15,
              minimum width=35mm, minimum height=35mm,
              below=5mm of mapping.south, anchor=north]
              (va-mppi) {};
        \node[font=\small] at (va-mppi.north) [yshift=-8pt] {Visibility-Aware MPPI};
        \draw [blue, thick, dotted, xshift=1.5cm, yshift=-3.5cm] plot [smooth, tension=1] coordinates { (0,0) (1.7,0.8) (2.2,1.5)};
        \draw [red, thick, dotted, xshift=1.5cm, yshift=-3.5cm] plot [smooth, tension=2] coordinates { (-0,0) (-0.7,1.2) (-0.5,2.3)};
        \draw [orange, thick, dotted, xshift=1.5cm, yshift=-3.5cm] plot [smooth, tension=2] coordinates { (-0,0) (0.4,1.2) (1.3,2.3)};
        \draw [violet, thick, dotted, xshift=1.5cm, yshift=-3.5cm] plot [smooth, tension=2] coordinates { (-0,0) (0.1,0.9) (-0.28, 1.8)};
        \pic[scale=1.0]   at (1.5,-1.5) {tree}; 
        \pic[scale=1.4]   at (2.5,-2.5) {tree};  
        \pic (perc) at (1.25, -3.25) {perception};
        \pic (robot) at (1,-3.5) {rzr};
        \node[module, below=10.5mm of mapping] (dyn) {Forward Sampling};
        \node[module, below=5mm of dyn] (unc) {Uncertainty Update};
        \node[module, below=5mm of unc] (mppi){MPPI Optimization};
        \draw[light, rounded corners=2pt] ([xshift=6.75mm]perc-north) -- ++(0, 33mm) -| ([xshift=-10mm]mapping.north) node[pos=0.3, below]{$\mathbf{y}_k$};
        \draw[flow] ([xshift=-10mm]mapping.south) -- node[right]{$(\boldsymbol{\mu}^e,\boldsymbol{\Sigma}^e)$} ([xshift=-10mm]va-mppi.north);
        \draw[flow] ([xshift=-10mm]dyn.south) -- node[right]{$\mathbf{X}^s$} ([xshift=-10mm]unc.north);
        \draw[flow] ([xshift=-10mm]unc.south) -- node[right]{$\{\boldsymbol{\Sigma}^e_1,\boldsymbol{\Sigma}^e_2\dots,\boldsymbol{\Sigma}^e_J\}$} ([xshift=-10mm]mppi.north);
        \draw[back, rounded corners=2pt]      
              ([xshift=-10mm]va-mppi.south) -- ++(0,-3mm) -| (robot-south) node[pos=.3, above] {$\mathbf{u}_k$}; 
        \end{tikzpicture}
    \caption{The overall architecture of the vision-aware MPPI controller, a vehicle operating in an unknown environment must build a stochastic map in real time. An uncertainty reduction mechanism is embedded in the prediction model allowing the controller to quantify uncertainty reduction along candidate trajectories(dotted lines).}
    \label{va_architecure}
\end{figure}
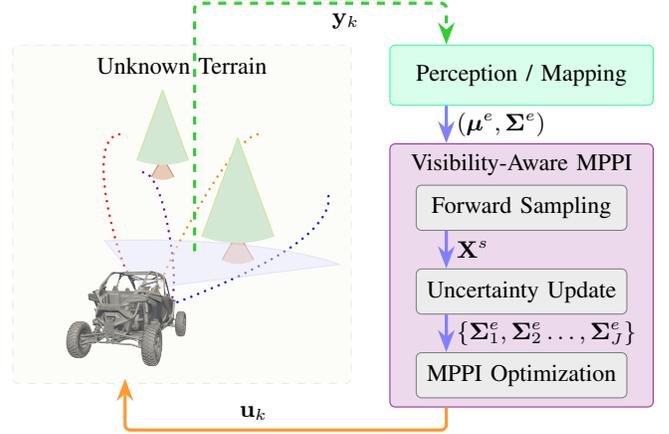

Here we present the formulation for the visibility-aware implicit-dual control problem. The overall system architecture is depicted in Fig~\ref{va_architecure}. Consider an off-road autonomous vehicle operating in a complex a priori unknown environment.  It builds a stochastic map of the environment in real time using onboard perception.  Due to the complex and cluttered nature of the environment, occlusion due to obstacles or large gradients are a common occurrence. Thus controlling the position of the vehicle directly affects the quality of the map. This highlights the dual nature of the problem: control actions influence the quality of the perception model, while the perception model, in turn, informs the optimal control strategy. We follow closely the notation used in~\cite{mesbah_pa} and~\cite{active_learning_dual} as these works are closest to our own in terms of objective. 

Let robot state, control input, and environment state at discrete time $k$ be denoted by $\mathbf{x}_k^s \in \mathbb{R}^{n_{x_s}}, \mathbf{u}_k \in \mathbb{R}^{m},$ and $\mathbf{x}_k^e \in \mathbb{R}^{n_{x_e}}$. The evolution of the robot's state is governed by the following discrete time dynamics:
\begin{equation}
\label{general_dynamics}
    \mathbf{x}^s_{k+1} = f^s(\mathbf{x}_k^s, \mathbf{u}_k, \mathbf{x}_k^e),
\end{equation}
where $f: \mathbb{R}^{n_x} \times \mathbb{R}^{m} \times \mathbb{R}^{n_{\uptheta}} \rightarrow \mathbb{R}^{n_{x}}$ is a (possibly nonlinear) state-transition function. It is subject to $\mathbf{x}^s \in \mathcal{X}$ and $\mathbf{u} \in \mathcal{U}$ where $\mathcal{X},\mathcal{U}$ represent the admissible sets for the robots state and control input respectively. Because the environment $\mathbf{x}^e$ is a priori unknown we model it online using onboard perception. We define the perception output function as
\begin{equation}
    \label{true_measurement}
    \mathbf{y}_k = h(\mathbf{x}_k^s, \mathbf{u}_k, \mathbf{x}_k^e),
\end{equation}
where $\mathbf{y}_k \in \mathbb{R}^{n_y}$ is the perception measurement vector at time step $k$. Eq.~(\ref{true_measurement}) encapsulates the dependence of the perception quality on the vehicles position within environment. Let $\mathbf{Y}_k = [\mathbf{y}_k, \dots, \mathbf{y}_0]$ represent the history of measurements up to time $k$. It contains all the information necessary to model $\mathbf{x}^e$. We define $\boldsymbol{\uptheta} \in \mathbb{R}^{n_{\uptheta}}$ as the parameter(s) needed to fully define the distribution of $\mathbf{x}^e$. That is, $\mathbf{x}_k^e  \sim b(\boldsymbol{\uptheta})$ where:
\begin{equation}
\label{probability}
b(\boldsymbol{\uptheta})= \text{P}[\boldsymbol{\uptheta}_k| \mathbf{Y}_k], 
\end{equation}
is the conditional probability of $\boldsymbol{\uptheta}_k$ given $\mathbf{Y}_k$. Eq.~(\ref{probability}) encodes the uncertainty in $\mathbf{x}^e$ which is updated recursively using Bayes' theorem through
\begin{subequations}
    \begin{equation}
    \label{measurement_step_true}
        \text{P}[\boldsymbol{\uptheta}_k|\mathbf{Y}_k] = \frac{\text{P}[\mathbf{y}_k|\boldsymbol{\uptheta}_k]\text{P}[\boldsymbol{\uptheta}_k|\mathbf{Y}_{k-1}]}{\text{P}[\mathbf{y}_k|\mathbf{Y}_{k-1}]},
    \end{equation}
    \begin{equation}
    \label{prediction_step_true}
        \text{P}[\boldsymbol{\uptheta}_{k+1}|\mathbf{Y}_k] = \int \text{P}[\boldsymbol{\uptheta}_{k+1}|\boldsymbol{\uptheta}_k, \mathbf{u}_k]\text{P}[\boldsymbol{\uptheta}_k|\mathbf{Y}_k]d\boldsymbol{\uptheta}_k.
    \end{equation}
\end{subequations}
Here Eqs.~(\ref{measurement_step_true}) and (\ref{prediction_step_true}) represent the correction step and prediction step of Bayesian estimation respectively. In this work, we model $\mathbf{x}^e$ as the elevation, with $\boldsymbol{\uptheta} := \mathcal{N}(\boldsymbol{\mu}^e, \boldsymbol{\Sigma}^e)$.
\begin{remark}
    In principle, $\boldsymbol{\uptheta}$ need not be Gaussian nor restricted to Bayesian methods—it may represent any probabilistic model of the environment.
\end{remark}
We define the sequences corresponding to the robot state, control inputs, and environment state over a prediction horizon of length $N \in \mathbb{Z}^+$ as follows:
\begin{align*}
    \mathbf{X}^s_k &= [\mathbf{x}_{0|k},\mathbf{x}_{1|k},\dots,\mathbf{x}_{N|k}], \\
    \mathbf{U}_k &= [\mathbf{u}_{0|k},\mathbf{u}_{1|k},\dots,\mathbf{u}_{N-1|k}], \\
    \mathbf{X}^e_k &= [\mathbf{x}^e_{0|k}, \mathbf{x}^e_{1|k}, \dots , \mathbf{x}^e_{N|k}].
\end{align*}
Here, notation $(i|k)$ denotes future step $i \in \mathbb{Z}^+$ given current time $k$. System performance in reaching a goal location is captured through minimizing a cost function defined over the prediction horizon as
\begin{equation}
    \label{expected_cost}
    \begin{aligned}
    J(\mathbf{X}_{k}^s, \mathbf{U}_{k}, \mathbf{X}_{k}^e) &= \mathop{\mathbb{E}}_{\mathbf{x}^e \sim b(\boldsymbol{\uptheta})_{0 \dots N|k}} [\phi (\mathbf{x}_{N|k}^s, \mathbf{x}^e_{N|k}) \\ 
    &+ \sum_{i=0}^{N-1}  l(\mathbf{x}_{i|k},\mathbf{u}_{i|k},\mathbf{x}^e_{i|k})],
    \end{aligned}
\end{equation}
where $\phi: \mathbb{R}^{n_s} \times \mathbb{R}^{n_e} \rightarrow \mathbb{R}$ and $l:\mathbb{R}^{n_s} \times \mathbb{R}^{m} \times \mathbb{R}^{n_e} \rightarrow \mathbb{R}$ are the terminal cost and stage cost respectively. 
\begin{remark}
    Eq.~(\ref{expected_cost}) computes the expectation of the cost at each time step in the prediction horizon with respect to the information available at that time step. This is critical for ensuring causality in the control decision~\cite{active_learning_dual}.
\end{remark}
While Eq.~(\ref{general_dynamics}) gives us a mechanism to predict future values of $\mathbf{x}^s$, Eqs~.(\ref{measurement_step_true}) and (\ref{prediction_step_true}) cannot directly provide future values of $\boldsymbol{\uptheta}$ due the reliance on future measurements $\mathbf{Y}_{i|k}$, which are unavailable. Consequently, we approximate the future distributions $b(\hat{\boldsymbol{\uptheta}}_{i|k})$ using predicted measurements $\hat{\mathbf{Y}}_{i|k}$. In the absence of a priori knowledge, predicting the future mean would be ill-posed. However, it can be confidently asserted that future measurements will reduce uncertainty. 

We define $b(\hat{\boldsymbol{\uptheta}})$ analogously to $b(\boldsymbol{\uptheta})$ as ${b}(\hat{\boldsymbol{\uptheta}}_{i|k}) = P[\hat{\boldsymbol{\uptheta}}|\hat{\mathbf{Y}}]$ where:
\begin{subequations}
    \begin{equation}
    \label{measurement_step_predicted}
        \text{P}[\hat{\boldsymbol{\uptheta}}_{i|k}|\hat{\mathbf{Y}}_{i|k}] = \frac{\text{P}[\hat{\mathbf{y}}_{i|k}|\hat{\boldsymbol{\uptheta}}_{i|k}]\text{P}[\hat{\boldsymbol{\uptheta}}_{i|k}|\hat{\mathbf{Y}}_{i-1|k}]}{\text{P}[\hat{\mathbf{y}}_{i|k}|\hat{\mathbf{Y}}_{i-1|k}]}
    \end{equation}
    \begin{equation}
    \label{prediction_step_predicted}
        \text{P}[\hat{\boldsymbol{\uptheta}}_{i+1|k}|\hat{\mathbf{Y}}_{i|k}] = \int \text{P}[\hat{\boldsymbol{\uptheta}}_{i+1|k}|\hat{\boldsymbol{\uptheta}}_{i|k}, \mathbf{u}_{i|k}]\text{P}[\hat{\boldsymbol{\uptheta}} _{i|k}|\hat{\mathbf{Y}}_{i|k}]d\hat{\boldsymbol{\uptheta}}_{i|k}.
    \end{equation}
\end{subequations}
Where
\begin{equation}
\label{predicted_measurment}
    \hat{\mathbf{y}}_{i|k} = \hat{h}(\mathbf{x}_{i|k}^s, \mathbf{u}_{i|k}, \mathbf{x}_{i|k}^e)
\end{equation}
is a predicted measurement vector. In this work, we employ an information-gain-driven approximation, treating future measurements as functions of the current mean estimate. The details of how Eqs.~(\ref{measurement_step_predicted}), (\ref{prediction_step_predicted}), and (\ref{predicted_measurment}) are implemented are outlined in Section~\ref{visibility_aware_uncertainty_update}. The sequence of predicted environment states is defined as
$$
    \hat{\mathbf{X}}_k^e = \{\hat{\mathbf{X}}_{0|k}^e, \hat{\mathbf{X}}_{1|k}^e, \dots, \hat{\mathbf{X}}_{N|k}^e \}.
$$
Finally, we pose the planning problem by combining Eqs.~(\ref{general_dynamics}), (\ref{expected_cost}), (\ref{measurement_step_predicted}), (\ref{prediction_step_predicted}), and (\ref{predicted_measurment}) in a stochastic optimal control formulation:
\begin{equation}
    \label{expected_optimal}
    \begin{aligned}
     \min_{\mathbf{U}} \quad &\mathop{\mathbb{E}}_{\hat{\mathbf{x}}^e\sim\hat{\boldsymbol{\uptheta}}_{N|k}} [\phi(\mathbf{x}_{N|k}^s, \hat{\mathbf{x}}_{N|k}^e] + 
     \sum_{i=0}^{N-1} \mathop{\mathbb{E}}_{\hat{\mathbf{x}}^e\sim\hat{\boldsymbol{\uptheta}}_{i|k}} [l(\mathbf{x}_{i|k}^s, \mathbf{u}_{i|k}, \hat{\mathbf{x}}_{i|k}^e)]  \\
   \text{s.t.}\quad & \mathbf{x}_{k+1}^s = f(\mathbf{x}_k^s, \mathbf{u}_k, \mathbf{x}_k^e), \quad i = 0, \dots, N-1, \\
    & \hat{\mathbf{x}}^e \sim \hat{b}(\hat{\boldsymbol{\uptheta}}_{i|k}), \quad i = 1, \dots, N \\
    & \hat{\mathbf{y}}_{i|k} = \hat{h}(\mathbf{x}_{i|k}^s, \mathbf{u}_{i|k}, \mathbf{x}_{i|k}^e), \quad k = 0, \dots, N-1 \\ 
    & \mathbf{x}_{0|k} = \mathbf{x}_k, \hat{\boldsymbol{\uptheta}}_{0|k} = \boldsymbol{\uptheta}_k \\
    & \mathbf{U}_k \in \mathcal{U}, \; \mathbf{X}_k \in \mathcal{X}.
\end{aligned}
\end{equation}
The formulation in Eq.~(\ref{expected_optimal}) establishes the dual control basis of the problem by integrating control and belief updates in unified framework. It implicitly accounts for environment uncertainties while balancing exploration and exploitation.
\section{Proposed Approach: Visibility-Aware Model Predictive Path Integral Control}
The proposed framework addresses navigation in unknown and cluttered environments by implicitly managing perception uncertainty through a dual control formulation. We introduce our visibility-aware uncertainty update mechanism first, as it underpins our overall approach.
\subsection{Visibility-Aware Uncertainty Update}
\label{visibility_aware_uncertainty_update}
\begin{figure*}[ht]
    \centering
    \hspace*{0cm}
    \includegraphics[width=1.0\textwidth]{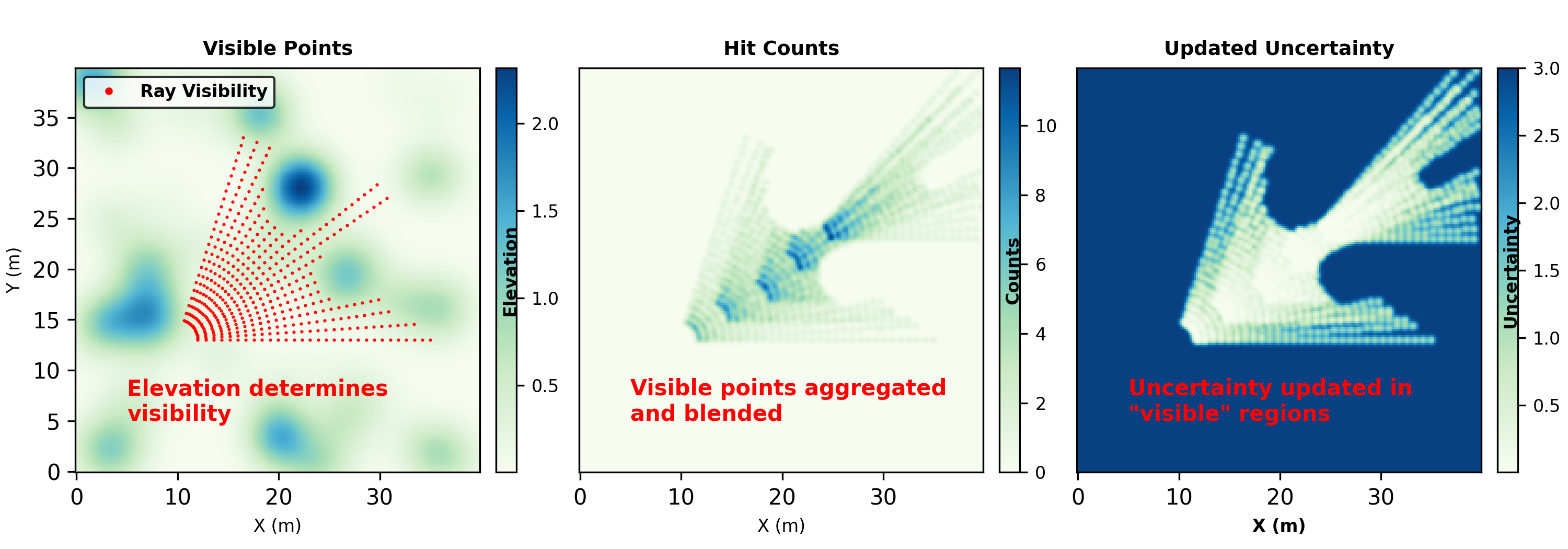}
    \caption{From left to right: 1. Rays are cast in a sparse grid over the mean elevation map and checked for visibility. 2. Visible points are aggregated along a candidate trajectory and added to a count map. A gaussian filter is applied to the final count map to spread the affect of each observed point. 3. The count map is convolved with the uncertainty map through a negative exponential function to reduce uncertainty.}
    \label{uncertainty_update}
\end{figure*}
One of the main differences between traditional stochastic controller formulation and dual-control formulation is the modeling of how the uncertainty will change along the prediction horizon. We use the predicted visibility of points along a candidate trajectory as a measure of uncertainty reduction. We assume there is direct correlation with the number of times a point is observed and the magnitude of the uncertainty reduction. 

To handle the predicted measurement of Eq.~(\ref{predicted_measurment}) we employ ray-casting over the mean elevation map $\boldsymbol{\mu}^e$. Fig.~(\ref{uncertainty_update}) depicts the steps employed to update the uncertainty along a candidate trajectory. We begin by initializing an observation count map $\mathcal{C}^j=0_{n_e}$ the same shape as $\boldsymbol{\mu}^e$ for each trajectory $\mathbf{X}_j^s$. As the vehicle moves along the trajectory, rays are cast in front of vehicle to determine visibility. For each visible point $\mathcal{C}^j$ is incremented by a count value $c$:
\begin{equation}
\label{count_increase}
   C_{x,y}^j = \begin{cases}
    c, \text{if point is visible} \\
    0, \text{otherwise},
\end{cases} 
\end{equation}
where subscript $(x,y)$ represent the $(X \& Y)$ coordinates of the point in the map frame.

\begin{figure}[h]
    \centering
    \hspace*{0cm}
    \includegraphics[width=1.0\linewidth]{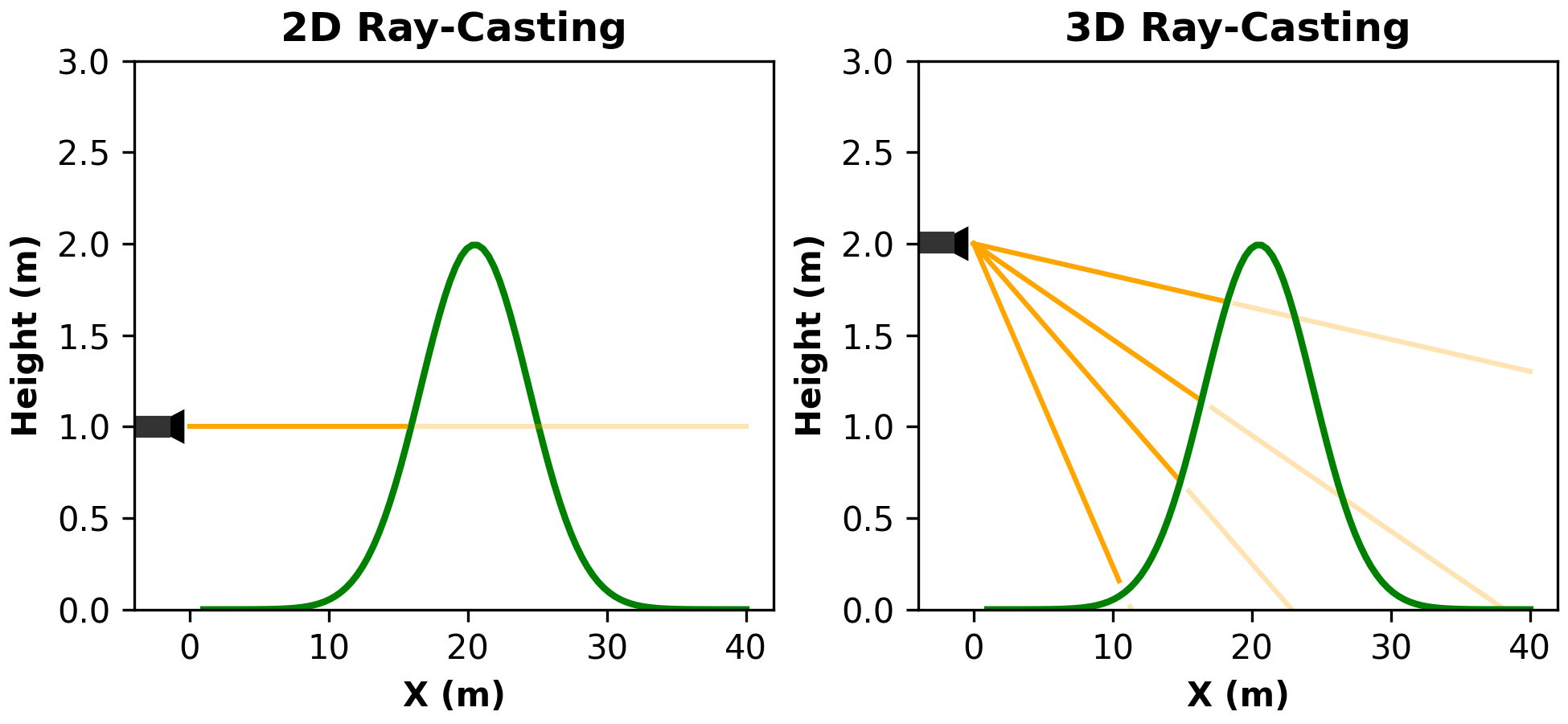}
    \caption{Rather than using dense 3D ray-casting in the visibility check, we approximate visibility with 2D rays and a threshold height value. The 3D model is still used to generate the ground truth map in simulations.}
    \label{flat_vs_3d_rays}
\end{figure}

Densely checking all points along a trajectory would be computationally prohibitive. We make two simplifications to significantly reduce the computation overhead. First, we approximate the true 3D visibility with a 2D horizontal ray and a threshhold visibility height $h$ as depicted in Fig.~\ref{flat_vs_3d_rays}.
Secondly, only a sparse set of points is checked for visibility (see left pane of Fig~\ref{uncertainty_update}). We use just 20 rays across the entire FOV$(72^{\circ}$) with just 30 equally spaced points along each ray. Each ray begins at 2(m) is limited to 25(m) which leads to checking visibility every 0.80(m). Such a sparse grid would do little to reduce uncertainty if we assume only the points observed contribute to the uncertainty reduction. To increase the effect of each visible point we "splat" it into neighboring cells with a Gaussian filter:
\begin{equation}
\label{gaussian}
    G(x, y) = \frac{1}{2\pi\sigma^2} \exp\left(-\frac{x^2 + y^2}{2\sigma^2}\right)
\end{equation},
where $G \in \mathbb{R}^{(9\times9)}$ and $\sigma = 1$. This gives an effective radius to each observation of about 0.9(m) with majority of the uncertainty reduction happening at the actual observed location. The count map is then convolved with the filter,
\begin{equation}
    \mathbf{C}^j = G \odot \mathcal{C}^j.
\end{equation}
The result is the count map gives the approximate number of times each point is observed along each trajectory.
\begin{remark}
    Due to the sparcity of the visibility check, we did find that occasionally we would get an "escaped ray" which did not intersect any point of the elevation map, see Fig.~\ref{uncertainty_update}. A denser visibility check would rectify this issue, but we did not find that it was worth the increased computational load.
\end{remark}    
\begin{figure}[h]
    \centering
    \hspace*{0cm}
    \includegraphics[width=1.0\linewidth]{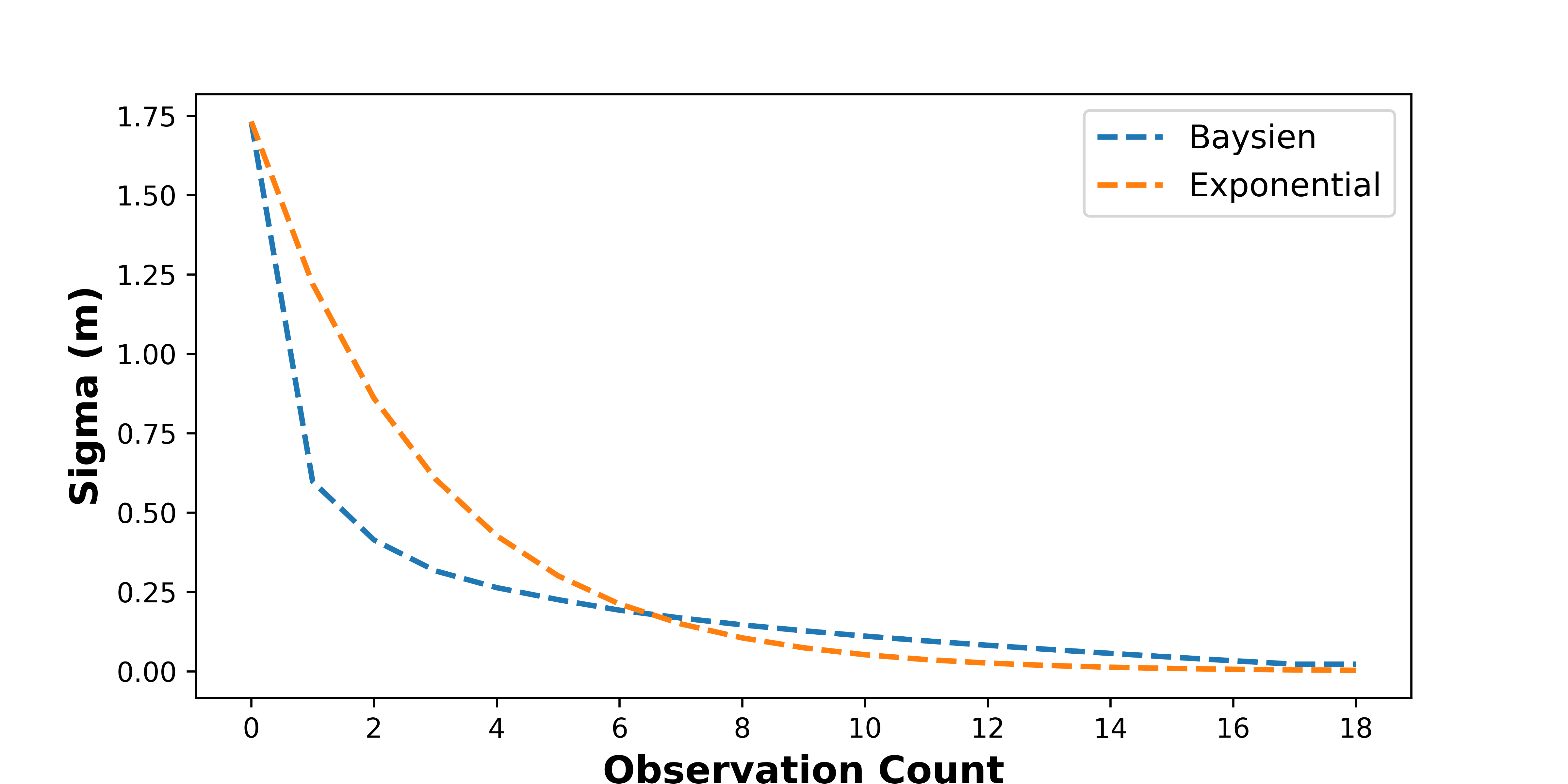}
    \caption{We approximate a true Bayesian update with an exponential function. Unlike Bayesian estimation, the exponential approximation can be performed in parallel improving real time performance. It also provides a "softer" decay forcing more observations for the same uncertainty reduction in the initial stages.}
    \label{bayes_update}
\end{figure}
The uncertainty $\boldsymbol{\Sigma}^e$ is updated according the number of times each point is observed. In a true Gaussian Bayesian estimation update there is a closed form solution:
$$
    \sigma^{n+1} = \frac{\sigma^n \cdot \sigma^m}{n \cdot \sigma^n + \sigma^m}
$$
where $\sigma_{new}, \sigma_0,$ and $\sigma_m$ are the updated, original and measurement uncertainties respectively and $n$ is the number of measurements taken. This update happens in a recursive manner as new measurements become available. However, because measurement uncertainties are dependent on the distance being measured, this would require updating the uncertainty map for each new set of measurements, which is computationally very expensive. Rather than applying a true Bayesian update, we approximate one using a negative exponential function:
 \begin{equation}
 \label{uncertainty_update_equation}
    \Sigma_k^j = \Sigma_k \odot \exp{(\gamma\cdot\mathbf{C}^j)} \approx \prod\frac{\Sigma_{x,y|k} \sigma_{x,y}^m}{\mathbf{C}_{x,y}^{j} \Sigma_{x,y|k} + \sigma_{x,y}^m},
 \end{equation}
 where $\sigma_{x,y}^m$ is the uncertainty for each range based measurement. 
The exponential gives a simple approximation which is easily tunable to adjust the aggressiveness of the uncertainty reduction and most importantly the update can performed in a parallel manner leading to faster uncertainty updates. Fig.~\ref{bayes_update} compares the uncertainty reduction vs number of observations for both a Bayesian and exponential approximation. While we did not implement it in this version, we recognize that adding some lowerbound to the uncertainty reduction would be an even better approximation of the sensor model which never truly converges to 0. The above simplifications allow for the visibility-aware uncertainty model to run in real time for up to 800 sample trajectories.
\begin{remark}
    By adjusting the value of $c$ in Eq.~(\ref{count_increase}) and $\gamma$ in Eq.~(\ref{uncertainty_update_equation})  we can change the aggressiveness of the uncertainty update, and thus vehicle behavior. Experiments showed that setting these values too low, prevented the vehicle from exploring, while too high decreased its safety. Similarly, the size of the Gaussian filter plays an important role. To small leaves large gaps in the count map, too large could wrongly assume non-visible points had been observed. 
\end{remark}

 \begin{remark}
     Because points are always projected in front of the vehicle, the uncertainty update always happens ahead of the vehicles location. On the other hand, when evaluating the cost function, the uncertainty only affects the vehicle at its current location. This is important for ensuring that causality is enforced in the control decision. As, while we update the uncertainty in a parallel approach after generating the rollouts, the uncertainty value along the trajectory will always be the result of past observations whether predicted or actual.
 \end{remark}
\subsection{Model Predictive Path Integral Control}
In implicit dual control problems, where uncertainty and control are tightly coupled, traditional optimization methods face significant challenges. The cost function depends not only on immediate performance objectives, but also on the evolution of uncertainty over time—a dependency that is both implicit and trajectory-specific. Solving such problems requires accurately estimating how control actions influence future states, sensor measurements, and ultimately the agent’s belief about the environment. Gradient-based methods are inherently ill-suited for capturing this trade-off, as they rely on smooth, well-behaved cost surfaces to compute reliable updates. In practice, the interplay between control and information acquisition introduces discontinuities that undermine gradient stability. Furthermore, trajectory optimization problems in complex environments are typically non-convex, compounding the difficulty of applying gradient-based solvers effectively.

The stochastic optimization problem in Eq.~\ref{expected_optimal}) formalizes this challenge: the uncertain parameter $\boldsymbol{\uptheta}$ is a function of the candidate trajectory, meaning the expectation must be evaluated over a trajectory-dependent distribution. This makes the true solution intractable, and necessitates the use of sampling-based approximations. In such settings, where expectations or chance constraints are estimated from a finite number of samples, gradient estimates become unreliable. Small perturbations in the control trajectory can lead to abrupt changes in sample validity, producing noisy or even vanishing gradients~\cite{streif2014stochasticnonlinearmodelpredictive}.

These difficulties are further exacerbated when the uncertainty model includes a visibility-based update, such as the ray-casting algorithm described earlier. Because this update must be recomputed for each candidate trajectory, it adds significant computational overhead. While not unique to our setting, this characteristic places additional demands on the control optimization method—particularly the need for efficient parallelization.

To simultaneously address the intractability of the expectation in Eq.~(\ref{expected_optimal}) and the non-convexity of the planning problem, we adopt the Model Predictive Path Integral (MPPI) framework~\cite{AggressiveMPPI}. MPPI circumvents the limitations of gradient-based methods by directly evaluating and weighting trajectories under stochastic dynamics. Its gradient-free formulation is naturally robust to non-smooth cost functions, and its sampling-based nature enables high levels of parallelism. These properties make MPPI a compelling choice for solving the implicit dual control problem under uncertainty.

Information Theoretic MPPI \textit{Williams et al}~\cite{williams2017informationtheoreticmodelpredictive} states that the optimal control policy $\mathbf{U}_k^*$ which minimizes Eq.~(\ref{expected_optimal}) can be represented by the optimal distribution $\mathbb{Q}^*$. By sufficiently sampling from this distribution, the mean control input will converge to the true optimal. However, we can not know $\mathbb{Q}^*$ directly. Rather we sample noise sequences $\boldsymbol{\varepsilon}_k^j=[\boldsymbol{\epsilon}_{0|k}^j,\boldsymbol{\epsilon}_{1|k}^j,\dots,\boldsymbol{\epsilon}_{N-1,k}^j]$ from a base distribution $\mathbb{P}$. These are added the current mean control sequence to give candidate control sequences $\mathbf{V}_k^j = [\mathbf{v}_{0|k}^j, \mathbf{v}_{1|k}^j, \dots, \mathbf{v}_{N-1|k}^j]$, that is $\mathbf{v}_{i|k}^j = \bar{\mathbf{u}}_{i|k}+\boldsymbol{\epsilon}_{i|k}^j$. This is akin to drawing samples from a controlled distribution $\mathbb{Q}_{\bar{\mathbf{U}},\Sigma_{\mathbf{u}}}$. Our goal then is to approximate $\mathbf{u}^*$ by driving $\mathbb{Q}$ ever closer to $\mathbb{Q}^*$, which is done by solving the following optimization:
\begin{equation}
\label{kldivergence}
    \mathbf{U}^* = \argmin_{\mathbf{U}} \mathbb{D}_{KL}(\mathbb{Q}^*||\mathbb{Q}),
\end{equation}
where operator $\mathbb{D}_{KL}$ is the Kullback-Leibler divergence. Eq.~(\ref{kldivergence}) can be rewritten as:
\begin{equation}
    \mathbf{U}^* = \mathop{\mathbb{E}}_{\mathbb{Q}} [w(V)\mathbf{V}_k]
\end{equation}
by following derivation in~\cite{williams2017informationtheoreticmodelpredictive}, where $w(V)$ is a weighting term determined by applying an exponential transform of the cost function:
\begin{equation}
    w^j(V) = \frac{\exp{-\frac{1}{\lambda}J(\mathbf{X}_k^s, \mathbf{V}^j, \mathbf{X}_k^e)}}{\sum_{j=1}^{J}\exp{-\frac{1}{\lambda}J(\mathbf{X}_k^s, \mathbf{V}^j, \mathbf{X}_k^e)}}.
\end{equation}
Finally this leads to the iterative control update law:
\begin{equation}
    \bar{\mathbf{U}}_{k+1} = \bar{\mathbf{U}}_{k} + \sum_{j=1}^{J}w^j(\mathbf{V})\cdot \boldsymbol{\varepsilon}_k^j
\end{equation}
\subsection{Implementation Details}
Here, we outline the implementation details of our visibility-aware MPPI controller. All components—vehicle dynamics, uncertainty updates, and visibility logic—are implemented in custom JAX kernels to support efficient parallelization and GPU execution.

Both the plant and prediction model employ a 3-DOF nonlinear bicycle model (Eq.~\ref{dynamics}) to predict vehicle behavior under high-speed off-road conditions.
\begin{equation}
\label{dynamics}
    f(X,U) = \begin{bmatrix}
       \frac{(f_{xt} - \frac{1}{2}\rho \cdot C_d \cdot A_f \cdot V^2 - m \cdot g \cdot C_r)}{m} + v \cdot r \\
       \frac{(-\sum_{i=1}^4 F_{yi})}{m} - V \cdot r \\
       \frac{(-lf \cdot (F_{11}+F_{12}) + lr \cdot (F_{21}+F_{22})}{Jx} \\
       V \cdot \cos{\Psi} - v \cdot \sin{\Psi}\\
       V \cdot \sin{\Psi} + v \cdot \cos{\Psi}\\
       r\\
    \end{bmatrix}
\end{equation}
\noindent where the state and control inputs are defined as:
$$
    X = [V, v, r, X, Y, \Psi]^T \& U = [\delta, f_ {xt}]^T.
$$
This model includes lateral and longitudinal load transfer, Ackerman steering, and a Pacejka nonlinear tire model, but neglects longitudinal slip, assuming the tractive force is applied at the center of mass. 
\begin{table}[ht]
    \centering
    \caption{Vehicle Parameters}
    \begin{tabular}{c l c}
        \toprule
        \textbf{Symbol} & \textbf{Description} & \textbf{Value} \\
        \midrule
        $m$ & mass & 1650kg \\
        $lf, lr$ & Front/Rear Vehicle length & 1.8(m) \\
        $tr$ & Track width & 2(m) \\
        $B,C,D,E$ & Pacejka parameters &  6, 2.5, 0.37, 1.1 \\
        $C_d$ & Aero dynamic drag coefficient & 0.7 \\
        $A_f$ & Frontal area & 4(m) \\
        $\rho$ & Air density & 1.225 \\
        $C_r$ & Longitudinal drag coefficient & 0.02 \\
        $g$ & gravity constant & 9.80655 \\
        \bottomrule
    \end{tabular}
    \label{veh_parameters}
\end{table}
For sensor modeling, we assume a forward-facing depth camera with a $72^{\circ}$ field of view and a maximum depth range of 25\,m. Visibility updates are computed using dense 3D ray-casting over a ground-truth map. While no noise is added to measurements, this is a reasonable assumption given the short range of the depth sensor. Uncertainty is updated using the exponential formulation described in Eq.~(\ref{uncertainty_update_equation}), though we omit Gaussian splatting since visibility is computed densely.

Unobserved space is assumed free (i.e. equal to robots elevation). However, occluded points within the field of view are nearest neighbor interpolated to complete the map. The uncertainty maps are initialized with no uncertainty near the robot and with 3.0 everywhere else. 
\subsection{Cost Function}
One key advantage of the MPPI framework is its ability to handle complex, non-convex, and discontinuous cost functions efficiently. Due to MPPI’s forward sampling approach, hard state constraints cannot be explicitly enforced. The proposed cost function prioritizes rapid and safe navigation to a designated goal, balancing multiple objectives including speed, vehicle stability, obstacle avoidance, and precise stopping criteria.
To incentivize high speed, we introduce a time-to-goal cost:
\begin{equation}
\label{time_cost}
    S_t = \omega_t(\frac{d_{20}}{4V_{20}} + \frac{d_{30}}{2V_{30}} + \frac{d_N}{V_N})
\end{equation}
where $d_*=\sqrt{(X_*-X_G)^2+(Y_*-Y_G)^2}$ represents the euclidean distance to the goal from specific points along the trajectory, and $V_*$ is the velocity at these points (at 2, 3, and 4 seconds ahead). We progressively increase the weight with terminal point being weighted the highest. We found this approach to give smoother and more consistent results than applying a terminal cost only. To ensure safe vehicle operation within dynamic limits, additional costs penalize excessive acceleration, side slip, rollover risk, and lateral velocity:
\begin{equation}
\begin{aligned}
S_{dynamic} = \omega_{dynamic} \sum_{i=0}^N(  \mathbb{I}(||\mathbf{a}_{y_i}|| \geq \mathbf{a}_{roll})+ \\
 \mathbb{I}(||\mathbf{a}_i|| \geq \mathbf{a}_{max}) + 
 \mathbb{I}(||\boldsymbol{\beta}_i|| \geq \boldsymbol{\beta}_{max}) + 
 \mathbb{I}(||\mathbf{v}_i|| \geq \mathbf{v}_{max}) ) , 
\end{aligned}
\end{equation}
where $\mathbb{I}$ is the indicator function defined as:
$$
    \mathbb{I} = \begin{cases}
        1, \text{if condition is true} \\
        0, \text{otherwise}.
    \end{cases}
$$
Here, ($\mathbf{a}, \mathbf{a}_y, \boldsymbol{\beta}, \& \  \mathbf{v}$) represent the total acceleration, lateral acceleration, side-slip and lateral velocity respectively. The limits ($\mathbf{a}_{max}$, and $\mathbf{a}_{roll}$) represent the traction limit and static stability factor of the vehicle, where as ($\boldsymbol{\beta}_{max}$ and $\mathbf{v}_{max}$) are user defined thresholds which can be adjusted to control aggressiveness of vehicle maneuvers. 

Obstacle avoidance is managed through elevation-based surface cost defined as the expectation of a sum of indicator functions evaluating multiple constraints along points projected into the vehicles footprint. Specifically, at each time step we project two arrays along the front and rear portion of the footprint which span the vehicles width. We then evaluate the vehicle x and y axis aligned gradients at these points. Additionally, we evaluate instantaneous step change across these arrays in addition to temporally along the prediction horizon. Finally, any point that is above a threshold elevation difference relative to the CG is also considered to be occupied. We evaluate 1000 samples of $x^e \sim b(\hat{\uptheta})$. The total surface cost is defined as:
\begin{equation}
\label{occupied}
\begin{aligned}
    &S_{surface} = \omega_{surface} \mathop{\mathbb{E}}_{\hat{\mathbf{x}}^e \sim b(\hat{\uptheta})} \sum_{i=0}^N\sum_{j=0}^{P-1} \Big[ \\
    &\mathbb{I} \left(\nabla_{X}^2 \mathbf{x}_{i,j}^e > 0.25 \right) + 
    \mathbb{I} \left(\nabla_{Y}^2 \mathbf{x}_{i,j}^e > 0.25 \right) + \\
    &\mathbb{I} \left(\Delta_{X}^2 \mathbf{x}_{i,j}^e > 0.1 \right) + 
    \mathbb{I} \left(\Delta_{Y}^2 \mathbf{x}_{i,j}^e > 0.1 \right) + \\
    &\mathbb{I} \left( \left\| \mathbf{x}_{i,j}^e - \mathbf{x}_0^e \right\| > 1.5 \right) \Big]
\end{aligned}
\end{equation}
where $\mathbf{x}_{i,j}^e$ represents one point in the projected arrays, $\nabla_{(.)}$ represents the gradient, and $\Delta_{(.)}$ is the instantaneous step change in elevation. 

These three costs are sufficient to navigate the vehicle toward a goal location quickly while avoiding dangerous behavior or potentially lethal obstacles. However, to encourage the vehicle to stop at the goal, we must also impose a set of costs which encourage approaching the goal slowly and in the desired orientation. The first of these is a distance cost:
\begin{equation}
\label{distance}
    S_{dist} = \omega_{dist}(\frac{d_N}{d_0} + 2\phi_G\frac{d_N}{d_0})
\end{equation}
where $d_N$ and $d_0$ are the terminal and initial distance to goal respectively. The term $\phi_G = 1-\frac{1}{1+\exp{(2d_0 + d_A})}$ is an activation function where $d_A$ is a tuning parameter and represents the distance at which the function begins to activate. Normalizing by the initial distance keeps the cost weighting relatively constant in the initial part of the trajectory, where as the activation function puts a much higher weight on distance as the vehicle nears the goal location acting as a potential well. This helped to eliminate unstable behavior of using time alone which would often overshoot the goal and need to come back around.

To encourage the vehicle to slow as it approaches the goal, we impose an acceleration based cost:
\begin{equation}
\label{stop}
    S_{acc} = \omega_{acc}(\frac{V_N^2||V_N||}{2d_N})
\end{equation}
where $\frac{V_N^2}{2d_N}$ is the constant acceleration value to bring the vehicle to exactly 0 velocity at the goal and $||V_N||$ is the terminal velocity and acts as an extra weighting factor penalizing excessive velocities. Finally, to ensure the vehicle arrives at the goal in the correct orientation, we impose a yaw angle cost:
\begin{equation}
\label{direction}
    S_{orientation} = \omega_{orientation}(\phi_G \sum_{i=0}^N ||\Psi_{i|k} - \Psi_G||\uppsi_i)
\end{equation}
where $\boldsymbol{\uppsi} = [0, \dots, 1]$ is a vector which progressively applies a higher weight along the prediction horizon.
\subsection{Comparison with baselines and evaluation metrics}
To evaluate the proposed visibility-aware MPPI approach comprehensively, we conduct comparisons with two baseline MPPI controllers: a prescient controller and a deterministic baseline, each emphasizing different aspects of uncertainty handling and knowledge of the environment. The prescient is a controller with a priori knowledge of the environment. It is myopic only in the sense that all predictive controllers are limited by their preview. While unrealistic, this controller represents the theoretical upper bound for performance for local trajectory planning. The deterministic is our baseline controller, like visibility-aware MPPI it must build the map in real time, but it plans only over the mean with no uncertainty sampling. 

To further validate the practical utility and robustness of our framework, we test our controllers extensively across two distinct simulation scenarios designed to challenge navigation under uncertainty. The first is a hypothetical scenario we call the "Alleyway" which represents a cluttered urban alleyway. The goal is placed at the end of the alley, with a large obstacle directly in the center. The vehicle must navigate around the obstacle, which occludes two other obstacles on either side.  Trajectories entering the unobserved space behind the first obstacle at too high velocity will not have time to slow the vehicle sufficiently to avoid the hidden obstacles. 

\begin{table}[ht]
    \centering
    \caption{Controller Parameters}
    \begin{tabular}{c l c}
        \toprule
        \textbf{Symbol} & \textbf{Description} & \textbf{Value} \\
        \midrule
        $J$ & Number MPPI samples & 400, 1000, 4000 \\
        $N_{\uptheta}$ & Number of uncertainty sampels & 1000 \\
        $N$ & Number of prediction steps & 40 \\
        $dt$ & discrete time step & 0.1(s) \\
        $x,y$ & Map length horizontal/vertical      & 80(m) \\
        $g$        & Grid resolution                & 0.2(m) \\
        fov & Field of view               & 72$^\circ$ \\
        $r$        & Ray length             & 25 m \\
        $N_\theta$ & Number of rays         & 20 \\
        $N_r$      & Number of ray points       & 30 \\
        $\gamma$   & Uncertainty decay coefficient  & 0.3 \\
        \bottomrule
    \end{tabular}
    \label{cont_parameters}
\end{table}

\begin{figure}[ht]
    \centering
    \hspace*{0cm}
    \includegraphics[width=1.0\linewidth]{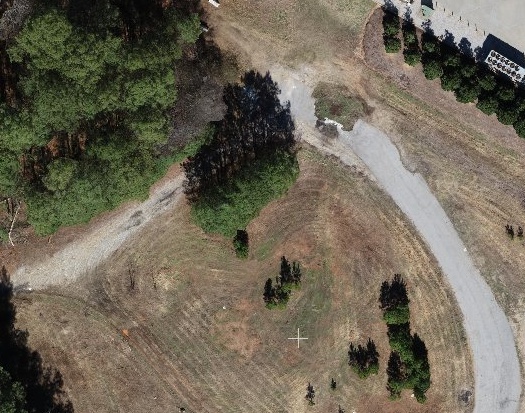}
    \caption{Aerial image showing the off-road environment we based our off-road simulation environment on.}
    \label{aerial}
\end{figure} 
The second scenario is based on an off-road test environment at our campus seen in Fig~\ref{aerial}. There is road running next to large open field, they are separated by a thin line of trees which occlude one from the other. The vehicle must reach goal location in the field from a starting point on the road. To do so it must navigate to the end of the row of trees and make a sharp left turn into the field. Each controller is evaluated on both scenarios at 400, 1000, and 4000 samples each. 
In order to be considered a successful mission the vehicle must:
\begin{itemize}
    \item Reach within 2(m) of the goal
    \item Be less than 1(m/s) within the goal region
    \item Not collide with any obstacles
\end{itemize}
A collision is an automatic failure, but a failure will also occur if not all stopping criteria are met. 
\section{Results and Discussion}
\subsection{Safety}
\begin{figure}[ht]
    \centering
    \hspace*{0cm}
    \includegraphics[width=1.0\linewidth]{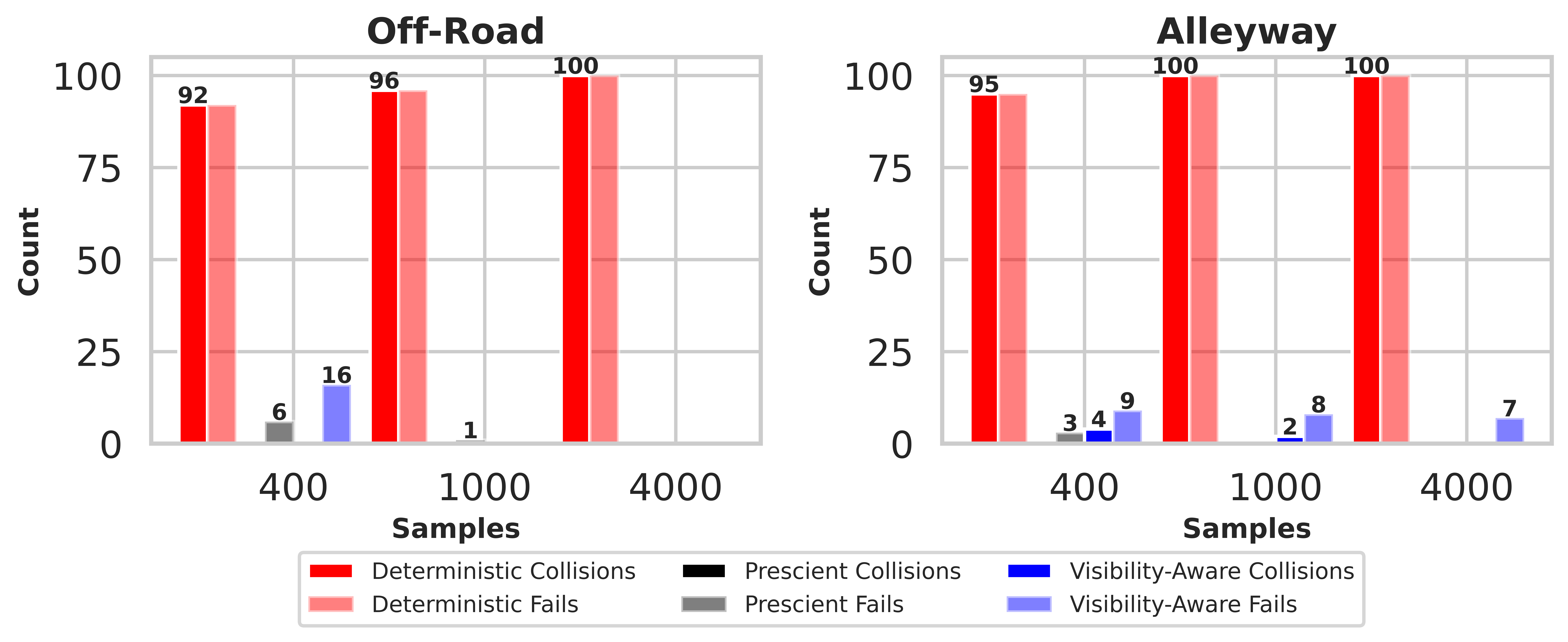}
    \caption{Shows the collisions and failures for both the off-road and alleyway environments.}
    \label{collisions}
\end{figure}
Each controller, scenario, and sample combination was repeated 100 times in order to generate statistically significant data. Fig.~\ref{collisions} summarizes the total number of collisions and failures for each of the scenarios. Visibility-aware MPPI shows a significant increase in safety over the deterministic case, while simultaneously giving up little performance as compared to the prescient case. In every scenario the deterministic controller is experiences far more failures than either the prescient or visibility-aware controller. In the 400 samples case for the off-road scenario, the deterministic controller had a 92\% failure rate, all of them due to collision with an unobserved obstacle. On the other hand the prescient and visibility-aware controllers had only 6\% and 16\% failure rate respectively none of them due to collision, rather they were the result of not being able to meet the stopping criteria. The Alleyway shows similar results with 95\%, 3\%, and 9\% failure rates respectively. Again all failures for the deterministic controller are due to collision. In this case 4 of the 9 total visibility-aware failures were the result of a collision. Another trend emerges from the plot that is somewhat counter-intuitive. As the number of samples increases, the number of failures for the deterministic controller also increase, in the 4000 sample scenarios there is a 100\% failure rate due to collision. This is a result of the assumption that unobserved space is free, and the controller finds increasingly aggressive maneuvers with more samples. Alternatively, the prescient and visibility-aware controllers see a decrease in the number of failures as samples increase. In the Alleyway, which is the arguable more difficult of the two scenarios, for 1000 and 4000 sample cases, the visibility-aware controller suffers 8\% and 7\% failure rates respectively. Of those, 2 of 8 and 0 of 7 respectively are due to collision.
\begin{figure}[h]
    \centering
    \hspace*{0cm}
    \includegraphics[width=1.0\linewidth]{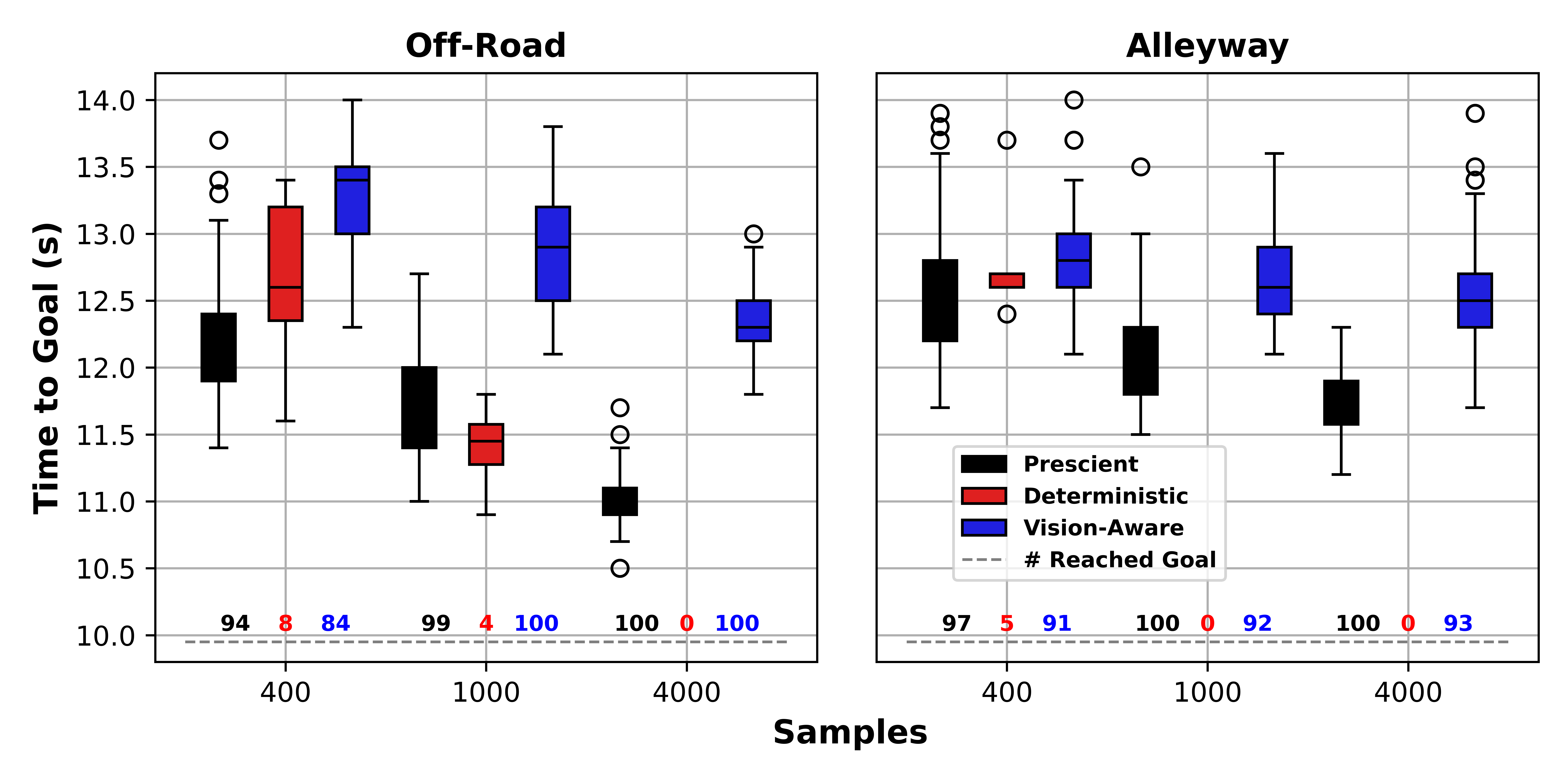}
    \caption{This plot displays the distribution of time to goal along with the number of successful runs for each controller. Time statistics are only plotted for successful missions, thus the deterministic controller shows no data in some scenarios.}
    \label{time_box}
\end{figure}
\subsection{Speed}
In spite of the additional safety, the visibility-aware controller yields very little in terms of performance. Fig.~\ref{time_box} displays the time to goal statistics for each of the test cases. In all cases, the for a given sample level, the prescient controller has the shortest time (except off-road-1000 samples, where deterministic had the shortest time) while the visibility-aware controller has the longest. In all cases, as the number of samples increased so too did the difference between the controllers. In the 400 sample cases the mean time for prescient and visibility aware controllers [12.2+/-0.5 (s), 13.3+/-0.4 (s)](off-road) and [12.5+/-0.5 (s), 12.8+/-0.4 (s)](Alleyway) which represents an increase of only 9.1\% and 2.5\% respectively. As the samples increase, the difference increases slightly to 12.2\% and 7.1\%. The increase is a result of the prescient controller finding better solutions, where as the visibility-aware controller also decreases time but retains the cautious approach.
\begin{figure*}[ht]
    \centering
    \hspace*{0cm}
    \includegraphics[width=1.0\textwidth]{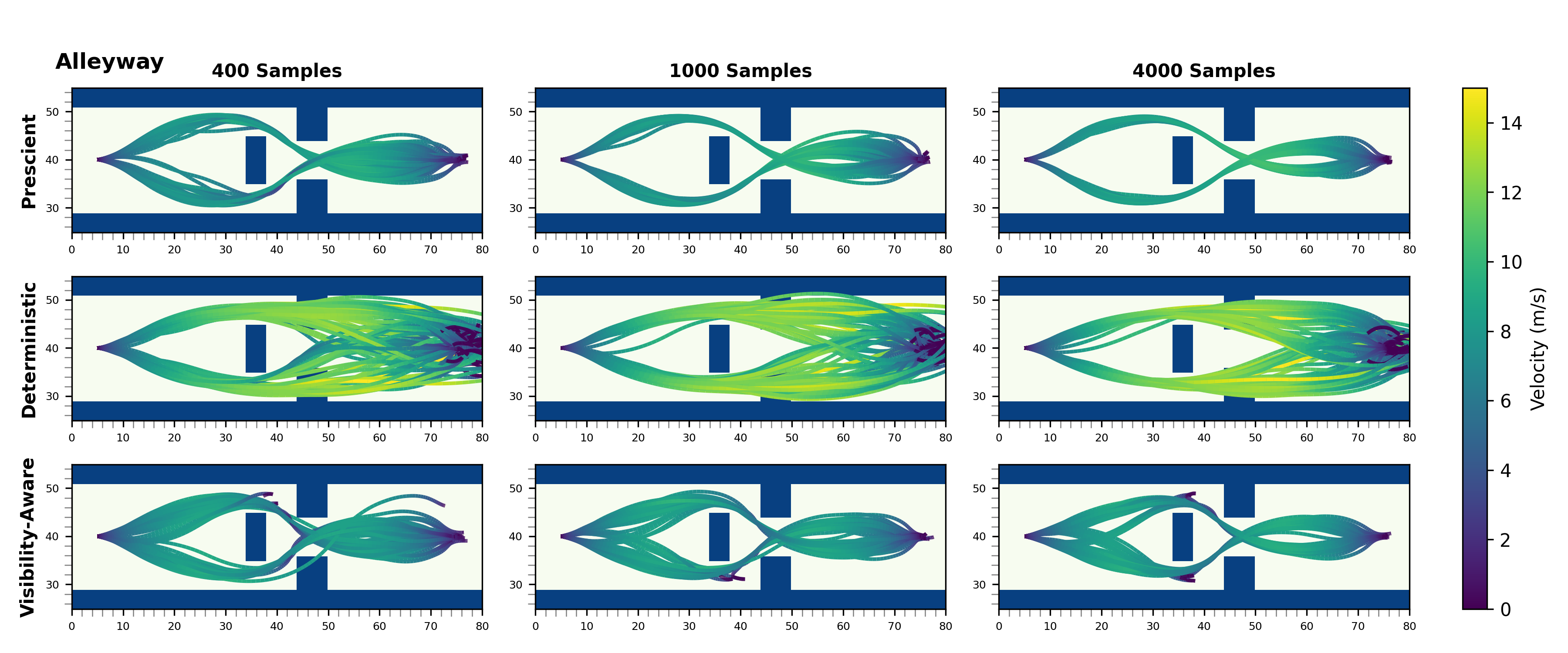}
    \caption{Trajectories for all simulations in the Corridor scenario. The deterministic controller (D-MPPI) maintains significantly higher speeds compared to the vision-aware and full-knowledge controllers. The vision-aware controller's trajectories closely resemble those of the full-knowledge controller, albeit slightly more conservative.}
    \label{alleyway_heatmap}
\end{figure*}
\begin{figure*}[ht]
    \centering
    \hspace*{0cm}
    \includegraphics[width=1.0\textwidth]{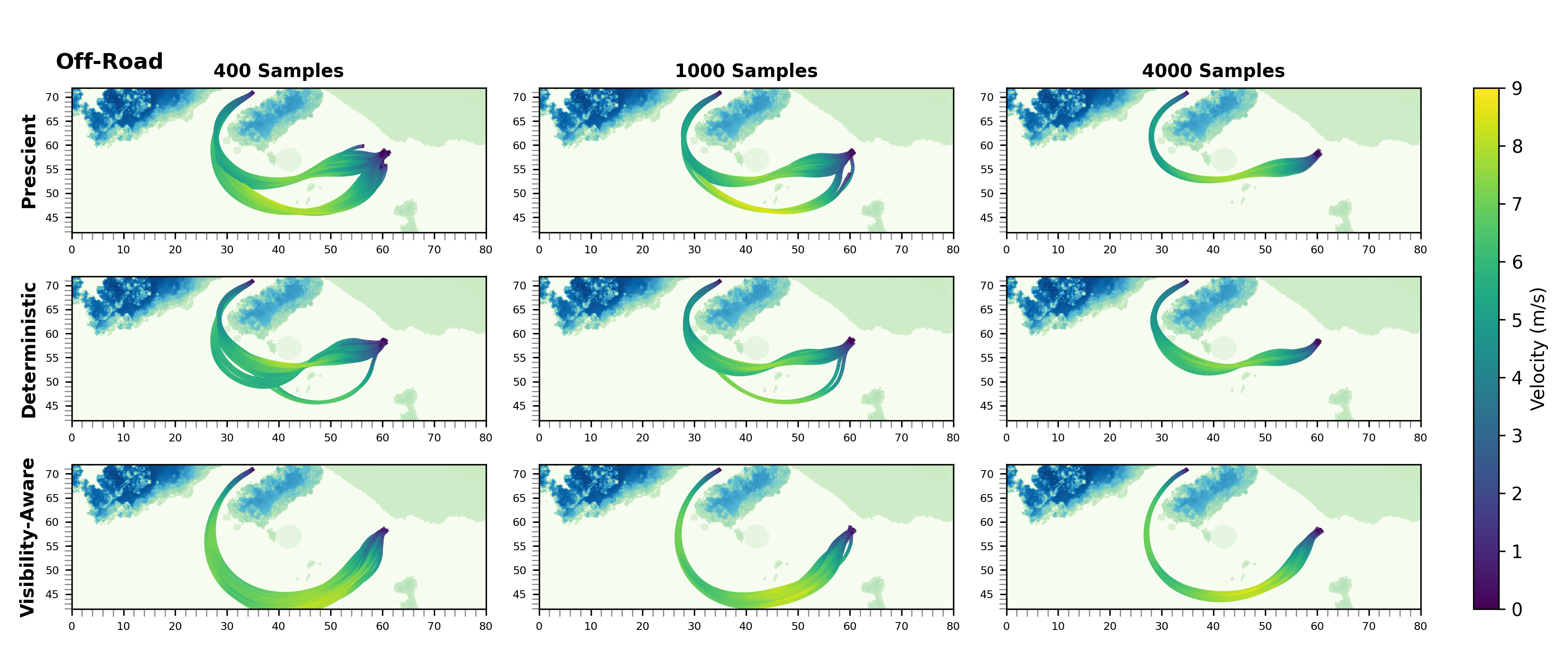}
    \caption{Trajectories for all simulations in the off-road scenario. The vision-aware controller navigates with greater caution, maintaining a wider trajectory into the unknown space. In contrast, the deterministic controller often turns prematurely into unobserved areas, resulting in frequent collisions with hidden obstacles.}
    \label{icar_heatmap}
\end{figure*}
The differing behavior the controllers evident in Fig.~\ref{alleyway_heatmap} and \ref{icar_heatmap} which show all trajectories (colored by velocity) for all simulations for Alleyway and off-road respectively. In the Alleyway scenario it is seen the deterministic controller exhibits much higher velocities in almost every test. This results in most trajectories experiencing a collision. There is also a significant spread in end position. The visibility-aware trajectories on the other hand exhibit lower average velocity trajectories which resemble the prescient controllers. Also, both prescient and visibility-aware are able to achieve much more consistent stopping locations at the goal. In the off-road scenario, another dangerous characteristic of the deterministic controller is revealed. Again, due to its assumption that unobserved space is free, it turns into the unobserved region of the field much earlier than the other two controllers, resulting (in most cases) in a collision with unobserved obstacles. The visibility-aware controller alternately takes a much safer and wider albeit longer path, which it makes up for with reaching higher velocities. 
\subsection{Visibility awareness}
\begin{figure}[ht]
    \centering
    \hspace*{-1cm}
    \includegraphics[width=1.2\linewidth]{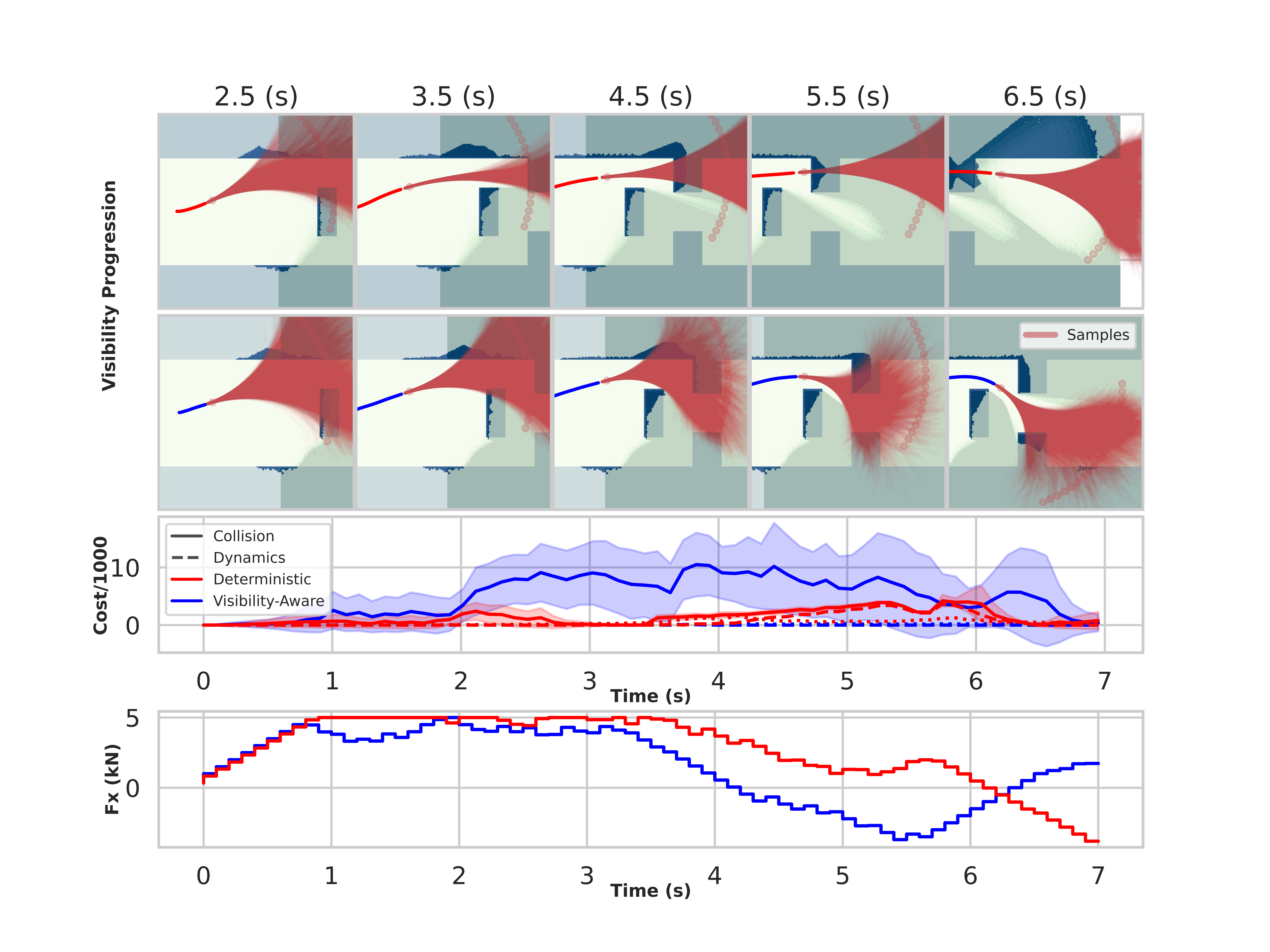}
    \caption{The top row shows the progression of visibility for the visibility-aware controller. The second row shows the surface cost. The last row shows drivetrain torque demand. Due to the free-space assumption, the deterministic controller carries a very high velocity into the unobserved space causing it to reach a point where it had no low collision cost options. This leads to a situation where it seeks to minimize other costs(such as time) and collision is imminent.}
    \label{visibility_progression}
\end{figure}

To better understand what leads to the increase in safety, the progression of visibility along with associated collision cost and control input was plotted for each both the deterministic and visibility-aware controllers. The results are shown in Fig.~\ref{visibility_progression}. It is noted that the deterministic controller does observe the gap between the obstacles, but due to its higher velocity, it has no trajectories which result in a safe path. This is seen in the cost function analysis in that up until 3.5(s) the minimum collision cost is reported as nearly 0. Once the second obstacle is observed the minimum cost begins to climb, implying no non-colliding trajectories are found. Therefore all paths will result in a collision and it simply tries to minimize other portions of the cost function(e.g. time) and increases the velocity further. The visibility-aware controller however, enters the unobserved space at a lower velocity which allows the samples to explore a wider region. This in turn allows for a safe path to be found and the controller successfully navigates the obstacles. 
\begin{figure}
    \centering
    \hspace*{0cm}
    \includegraphics[width=1.0\linewidth]{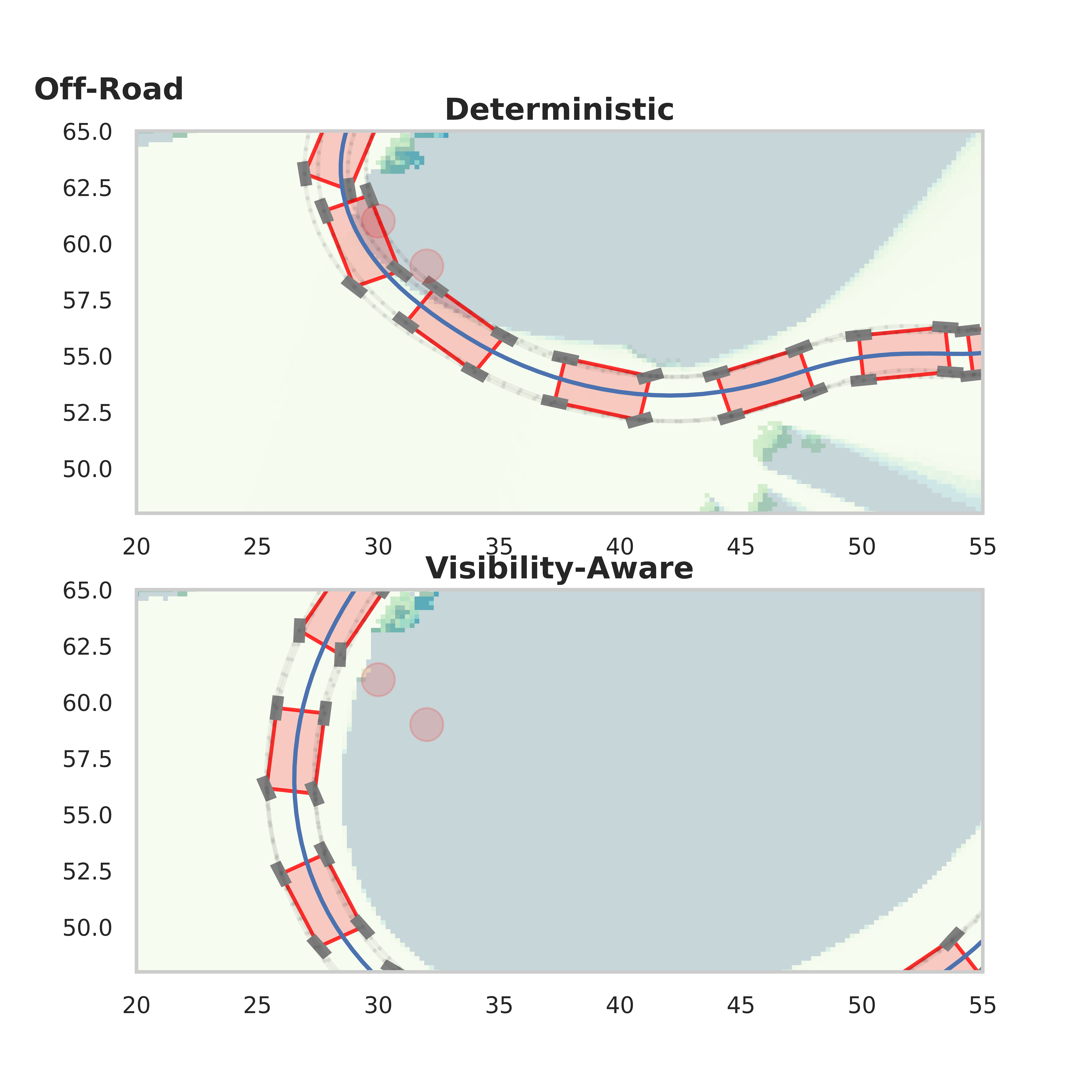}
    \caption{In scenario two we see that neither of the vision-aware or deterministic controllers actually observed the obstacle. But, due to the high uncertainty associated with the unobserved space, the vision-aware controller turns in less aggressively thus avoiding the obstacle.}
    \label{unobserved}
\end{figure}
Fig.~\ref{unobserved} depicts another scenario which reveals the differing behavior of the deterministic and visibility-aware controllers. The plot shows the actual map built by each of the controllers respectively. Observed space is represented by green-blue colors and the unobserved (and thus uncertain) space is gray. Two unobserved obstacles are shown as red dots. In this case neither controllers observed the hidden obstacles. However, the deterministic controller turns into this unobserved space resulting in collision with the obstacles. While the visibility-aware controller takes wider turn and never enters unobserved space. We do not explicitly ask the controller to avoid unobserved space, rather this happens implicitly as result of its associated uncertainty.
Because our controller lacks the explicit positional based visibility or information gain metrics, we often to do not observe the traditional wide curves around occluding corners. However, due to the dual control formulation, it has an implicit mechanism by which it can reduce the uncertainty allowing it to explore (albeit cautiously) unexplored space.

\begin{figure}[ht]
    \centering
    \includegraphics[width=\linewidth]{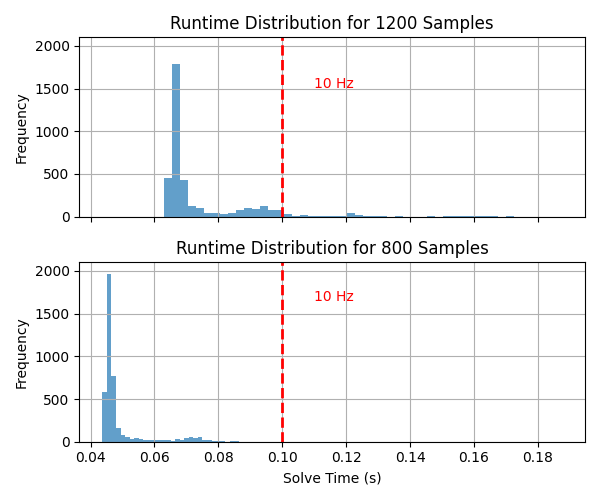}
    \caption{Distributions of solver runtimes (Top: 1200 samples, Bottom: 800 samples). The mean solve time for 1200 samples is below the 10(Hz) control frequency, but a few points exceed the limit. However, at 800 samples the controller always solves within the time limit.}
    \label{solvetime}
\end{figure}
\subsection{Real Time Performance}
Finally, we assess the real-time applicability of our visibility-aware MPPI through a realistic ROS-based simulation setup. Given the computational demands of both trajectory sampling and uncertainty updates, real-time validation is crucial for demonstrating practical deployment potential. We selected the alley way environment for this experiment. To emulate the depth perception, we used a custom OptiX ray tracing model to generate a dense point cloud at high speed. We used the GPU accelerated elevation mapping algorithm from~\cite{mapping} to return the mean elevation map (inpainted with Telea algorithm) and variance with all unseen areas initialized to 1(m) variance. Visibility-Aware MPPI is a custom JAX implementation and was set to 800 samples per time step. The prediction model given by Eq.~\ref{dynamics}, was set to 10(Hz) with a 4 second preview horizon. The plant model was running in its own node and used the same model as the prediction model, but set to run at 50(Hz). The whole pipeline was deployed on an Intel Xeon w9-3495X 56 core CPU with 1 TB Ram and an NVIDIA A6000 ADA series GPU. A histogram of the controller solve time can be seen in Fig.~\ref{solvetime} Controller solve time was 50 ms +/- 10 ms with the vast majority of the time going to the visibility update. 
\section{Conclusion and Future Work}
This work addresses the critical challenge of navigating complex, cluttered, and unstructured environments that are a priori unknown, with a particular focus on incorporating perception uncertainties into the planning and control process. A novel vision-aware Model Predictive Path Integral (VA-MPPI) framework was developed, leveraging an implicit dual-control formulation to naturally balance exploration and exploitation. By reasoning over vision uncertainties without the need for explicit perception objectives, the proposed framework enables robust navigation in highly uncertain and occluded environments, such as off-road and off-trail scenarios.

Simulation results demonstrate that the VA-MPPI controller significantly outperforms the deterministic baseline in terms of safety, reducing collision rates across all tested scenarios. Notably, the controller achieved this improvement while yielding only a marginal increase in time to goal compared to the prescient controller, which represents the theoretical upper bound of performance. These results highlight the controller's ability to implicitly handle perception uncertainties, avoid unobserved obstacles, and adaptively explore unknown environments.

Qualitative analysis of the trajectories further revealed that the deterministic controller's assumption of free unobserved space leads to aggressive and unsafe behavior, while the VA-MPPI controller adopts a cautious approach that implicitly prioritizes safety without overly compromising performance. The progression of visibility and the associated uncertainty updates showed how the VA-MPPI controller successfully integrates uncertainty reduction into the control pipeline, enabling it to find feasible and safe paths through occluded regions.

Despite these advances, the VA-MPPI controller exhibited a small number of failures in specific high-sample scenarios, largely due to limitations in the prediction model at very low velocities. Future work could address these limitations by integrating a more robust or adaptive prediction model to further improve reliability in extreme cases.

Other potential areas of improvement include,
\begin{itemize}
    \item Decreasing computational demands through more intelligent visibility checking
    \item Incorporating semantic scene understanding in addition to the geometric representation
    \item Incorporating a higher fidelity prediction model which can account for vehicle terrain interaction through geometric and dynamic consderations: i.e. suspension dynamics, wheel slip, etc..
\end{itemize}
In conclusion, the VA-MPPI framework represents a significant step forward in enabling autonomous ground vehicles to navigate unknown, occluded environments by integrating perception uncertainties directly into the planning process. This implicit dual-control approach provides a powerful foundation for safe and efficient navigation, paving the way for future research into dynamic, perception-aware navigation systems for real-world applications.
\section{Acknowledgement}
This work was supported by the Virtual Prototyping of Autonomy Enabled Ground Systems (VIPR-GS), a US Army Center of Excellence for modeling and simulation of ground vehicles, under Cooperative Agreement W56HZV-21-2-0001 with the US Army DEVCOM Ground Vehicle Systems Center (GVSC). DISTRIBUTION STATEMENT A. Approved for public release; distribution is unlimited. OPSEC9799
\FloatBarrier

\bibliographystyle{ieeetran}
\bibliography{refs}
\vspace{-1.0cm} 
\begin{IEEEbiography}[{\includegraphics[width=1in,height=1.25in,clip,keepaspectratio]{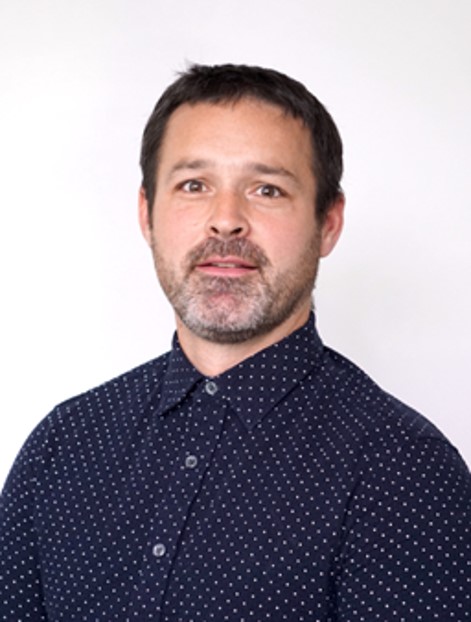}}]{Benjamin Johnson}
received his B.S. in Mechanical Engineering from University of Washington, his M.S. in Automotive Engineering from Clemson University and is currently a Ph.D. candidate in Automotive Engineering at Clemson University. His research interests include off-road navigation, uncertainty-aware planning, and autonomous vehicle control.
\end{IEEEbiography}

\begin{IEEEbiography}[{\includegraphics[width=1in,height=1.25in,clip,keepaspectratio]{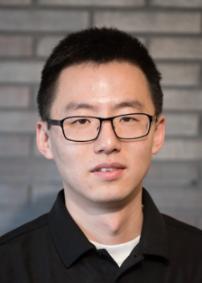}}]{Qilun Zhu}
received a B.S degree from Jilin University, Changchun, Jilin, China, in 2010, M.Eng from Cornell University, Ithaca, NY, USA, in 2011, and a Ph.D. degree from Clemson University, Greenville, SC, USA, in 2015. 
Dr. Zhu is currently working as an Assistant Research Professor in the Department of Automotive Engineering at Clemson University. He was also appointed as a Fellow of the Robert H. Brooks Sports Science Institute. His research interests include theories and applications of estimation, identification, learning, and optimal control. Dr. Zhu is a member of the Automotive and Transportation Systems Technical Committee and Automotive Controls Technical Committee. He serves as associate editor of Proceedings of the IMechE, Part D: Journal of Automobile Engineering and Conference on Control Technology and Applications. Dr. Zhu received awards of the SAE Young Professional Technical Paper Competition and Best Paper of 2014 ACC. 
\end{IEEEbiography}

\begin{IEEEbiography}[{\includegraphics[width=1in,height=1.25in,clip,keepaspectratio]{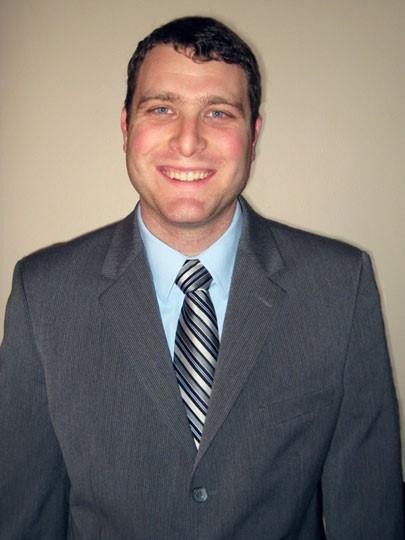}}]{Robert Prucka}
received the B.S.E degree, M.S.E degree and Ph.D. degree in mechanical engineering from University of Michigan, Ann Arbor, Michigan, USA in 2000, 2004 and 2008 respectively.  
    He is a Professor in the Department of Automotive Engineering at the Clemson University – International Center for Automotive Research. His research and teaching interests include the design, performance, control, calibration, and emissions of advanced internal combustion engines. Dr. Prucka is also the team leader for one of the current Deep Orange prototype vehicle programs at Clemson University, the faculty advisor for the Clemson University – Racing student team and is active in the motorsports engineering initiatives of the Brooks Sports Science Institute. Dr. Prucka was the recipient of Clemson University’s Murray Sokely Award in 2015, recognizing his outstanding contributions to engineering education.
\end{IEEEbiography}

\begin{IEEEbiography}[{\includegraphics[width=1in,height=1.25in,clip,keepaspectratio]{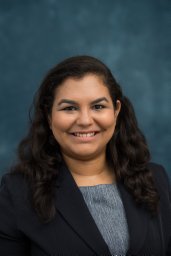}}]{Miriam Figueroa-Santos}
received a B.S. degree in mechanical engineering from the University of Puerto Rico, Mayaguez Campus, USA in 2015. She received her M.S.E. and Ph.D degree in mechanical engineering from University of Michigan, Ann Arbor, Michigan in 2018 and 2021 respectively. She is currently working at the U.S. Army DEVCOM Ground Vehicle Systems Center (GVSC) as a member of the Vehicle Performance and Analytic Development M\&S team. Dr. Figueroa-Santos provides analytical support on future tactical vehicle concepts.
\end{IEEEbiography}

\begin{IEEEbiography}[{\includegraphics[width=1in,height=1.25in,clip,keepaspectratio]{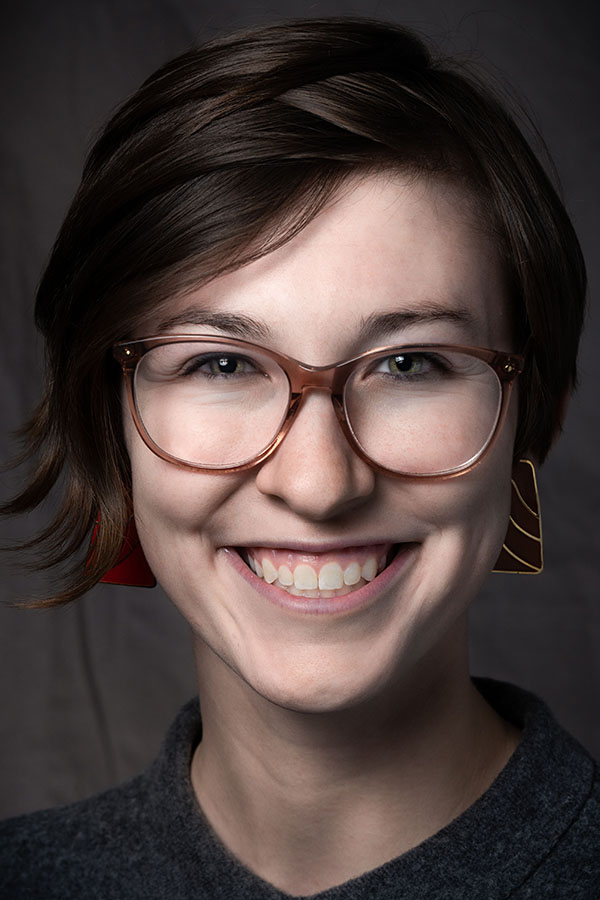}}]{Morgan Barron}
earned her B.S. degree in mechanical engineering from the University of Utah, Salt Lake City, Utah in 2020 and is currently working towards a M.S. degree in systems engineering at the University of Michigan, Ann Arbor.  She is a member of the Vehicle Performance and Analytic Development M\&S team at U.S. Army DEVCOM Ground Vehicle Systems Center (GVSC). Ms. Barron specializes in next generation military vehicle powertrain modeling.
\end{IEEEbiography}

\begin{IEEEbiography}[{\includegraphics[width=1in,height=1.25in,clip,keepaspectratio]{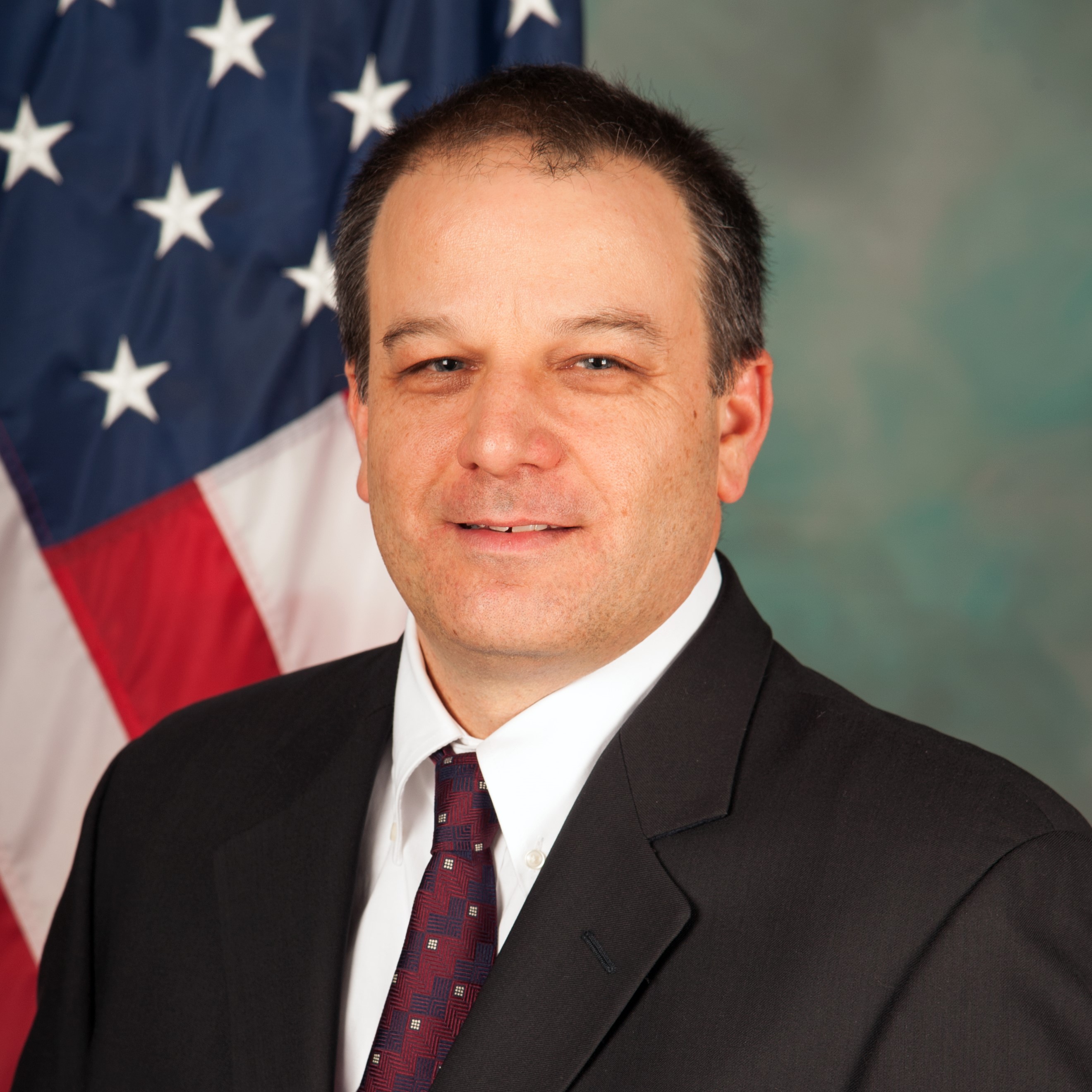}}]{Matthew P. Castanier}
is a Research Mechanical Engineer at the US Army DEVCOM Ground Vehicle Systems Center (GVSC) in Warren, MI. His research interests include powertrain modeling and simulation, structural dynamics analysis, and trade space exploration. He received his Ph.D. in Mechanical Engineering from the University of Michigan (UM) in 1995. From 1996 through 2008 he was a member of the research faculty at UM. He joined GVSC in 2008 and was promoted to his current position in 2011. Dr. Castanier has published more than 50 journal articles and over 100 conference papers.
\end{IEEEbiography}

\end{document}